\documentclass[10pt,a4paper]{article}

\usepackage[utf8]{inputenc}
\usepackage[T1]{fontenc}
\usepackage{hyperref}
\usepackage{url}
\usepackage{booktabs}
\usepackage{amsfonts}
\usepackage{nicefrac}
\usepackage{microtype}
\usepackage{xcolor}
\usepackage{subcaption}
\usepackage{makecell}

\usepackage{graphicx}

\usepackage{amsmath,amssymb,amsthm}
\usepackage{bm,fixmath}

\DeclareMathOperator*{\argmin}{arg\,min}

\DeclareMathOperator{\prox}{prox}

\DeclareMathOperator{\proj}{proj}

\newtheorem{theorem}{Theorem}[section]

\newtheorem{lemma}[theorem]{Lemma}

\newcommand{\norm}[1]{\left\lVert#1\right\rVert}

\newcommand{\N}{\mathbb{N}}
\newcommand{\R}{\mathbb{R}}
\newcommand{\f}{\mathbold{f}}
\newcommand{\bdelta}{\mathbold{\delta}}
\newcommand{\transp}{\top}
\newcommand{\av}{\mathbold{a}}
\newcommand{\bv}{\mathbold{b}}
\newcommand{\e}{\mathbold{e}}

\newcommand{\q}{\mathbold{q}}
\newcommand{\tv}{\mathbold{t}}
\newcommand{\x}{\mathbold{x}}
\newcommand{\y}{\mathbold{y}}
\newcommand{\vv}{\mathbold{v}}
\newcommand{\w}{\mathbold{w}}
\newcommand{\z}{\mathbold{z}}
\newcommand{\xx}{\pmb{\xi}}
\newcommand{\ball}[2]{\mathbb{B}_{#2}(#1)}
\newcommand{\Am}{\mathbold{A}}
\renewcommand{\Im}{\mathbold{I}}
\newcommand{\Pm}{\mathbold{P}}
\newcommand{\cv}{\mathbold{c}}
\newcommand{\I}{\mathcal{I}}
\newcommand{\T}{\mathcal{T}}

\makeatletter
\let\@fnsymbol\@arabic
\makeatother

\begin{document}

\title{OpReg-Boost: Learning to Accelerate Online Algorithms with Operator Regression}

\author{Nicola Bastianello%
\thanks{N. Bastianello is with the Department of Information Engineering (DEI), University of Padova, Italy. {\tt\small nicola.bastianello.3@phd.unipd.it}}, \

Andrea Simonetto%
\thanks{A. Simonetto is with UMA, ENSTA Paris, Institut Polytechnique de Paris, France. {\tt\small andrea.simonetto@ensta-paris.fr}}, \

Emiliano Dall'Anese%
\thanks{E. Dall'Anese is with the Department of Electrical, Computer, and Energy Engineering, University of Colorado Boulder, Boulder, Colorado, USA {\tt\small emiliano.dallanese@colorado.edu}}
}

\maketitle

\begin{abstract}
This paper presents a new regularization approach -- termed \emph{OpReg-Boost} -- to boost the convergence of online optimization and learning algorithms. In particular, the paper considers online  algorithms for optimization problems with a time-varying (weakly) convex composite cost. For a given online algorithm, OpReg-Boost learns the closest algorithmic map that yields linear convergence; to this end, the learning procedure hinges on the concept of \emph{operator regression}. We show how to formalize the operator regression problem and propose a computationally-efficient Peaceman-Rachford solver that exploits a closed-form solution of simple quadratically-constrained quadratic programs (QCQPs). Simulation results showcase the superior properties of OpReg-Boost w.r.t. the more classical forward-backward algorithm, FISTA, and Anderson acceleration.
\end{abstract}

\section{Introduction}\label{sec:introduction}

In recent years, the increasing volume of streaming data in many engineering and science domains has stimulated a growing number of research efforts on online optimization and learning (\cite{popkov2005gradient,Besbes2013,Asif2014,hall2015online,Jadbabaie2015, mokhtari2016online,SPM,NaLi2020} and many others). In data processing and machine learning applications, the cost function and the constraints (if present) are parametrized over data points that arrive sequentially; consequently, cost and constraint are time-dependent to reflect new data points and possibly time-varying learning objectives. {Beyond data processing and machine learning applications, emerging problems in the context of learning-based control have stimulated lines of research in online  identification of dynamical systems \cite{zheng_nonasymptotic_2021}, and online optimization for robotics~\cite{Berkenkamp2016,Luo2020}, model predictive control~\cite{paternain_prediction_2018,liaomcpherson_semismooth_2018,zhang_regret_2021}, and games \cite{Bel2021,fabiani_learning_2021}, to name a few.}

Let now $k \in \N$ and $F_k(\x)$ be a time-varying function, then formally we are interested in time-varying problems of the form
\begin{equation}\label{eq:base-problem}
	\x_k^* \in \argmin_{\x \in \R^n}  F_k(\x) := f_k(\x) + g_k(\x) 
\end{equation}
In particular, we assume that $f_k : \R^n \to \R$ is closed, proper, and $\mu$-weakly convex\footnote{
{\bf Notation}. We say that a function $f:\R^n \to \R$ is $\mu$-weakly convex if $f(x) + \mu/2\|\x-\x_0\|^2_2$, with $\mu >0$, is convex. The set of convex functions on $\R^n$ that are $L$-smooth (\emph{i.e.}, have $L$-Lipschitz continuous gradient) and $\mu$-strongly convex is denoted as $\mathcal{S}_{\mu,L}(\R^n)$, for $\mu, L > 0$; $\mathcal{S}_{0,L}(\R^n)$ is the set of $L$-smooth  convex functions. %
An operator $\T: \R^n \to \R^n$ is non-expansive iff $\|\T(\x) - \T(\y) \| \leq \|\x- \y\|$, for all $\x, \y \in \R^n$; on the other hand, $\T: \R^n \to \R^n$ is $\zeta$-contractive, with $\zeta\in(0,1)$, iff $\|\T(\x) - \T(\y) \| \leq \zeta \|\x- \y\|$, for all $\x, \y \in \R^n$. We denote the composition of two operators $\T_1, \T_2$ as $(\T_1 \circ \T_2)(\x) = \T_1(\T_2(\x))$. We denote by $\I$ the identity map $\I(\x) = \x$.
We define as $\prox_{\alpha g}(\y) = \argmin_\x \left\{ g(\x) + \norm{\x - \y}^2 / (2\alpha) \right\}$ the proximal operator of a function $g$ with parameter $\alpha>0$, and we denote by $\proj_{C}(\cdot)$, the projection operator onto the set $C$.
We denote by $\Im_n$ the identity matrix of size $n$, and by $\otimes$ the Kronecker product.
} for each $k \in \N$, and $g_k: \R^n \to \R \cup \{ +\infty \}$ is closed, convex and proper uniformly in time (optionally, one can also consider a setting where $g_k \equiv 0$). The goal is to design an \emph{online algorithm} $\mathcal{A}_k: \R^n \rightarrow \R^n$, with updates $\x_{k} = \mathcal{A}_k(\x_{k-1})$, so that the sequence  $\{\x_k\}_{k\in \N}$ exhibits an asymptotic behavior $\limsup_{k \to \infty}  F_k(\x_k) - F_k^* \leq B < \infty$, for a properly defined sequence of optimal value functions $\{F_k^*\}_{k \in \N}$ and with $B$ as small as possible. For this result to be feasible, a blanket assumption common in the online optimization literature is that the variations of problem in time (in terms of path length or functional variability) can be upper bounded by a sub-linear or a linear function of $k$; see \emph{e.g.},~\cite{Besbes2013, Jadbabaie2015, SPM, mokhtari2016online,NaLi2020,hallak2020regret} . If this latter function is linear in $k$, then it is known that online algorithms  exhibit an asymptotic error.

A key intuition is to use the existence of this error as an advantage: given the presence of an error due to the dynamics of the cost, one can leverage regularizations in the optimization problem or modifications of the algorithmic steps to boost the convergence without necessarily sacrificing performance. Surprisingly, \emph{there may be no trade-off between accuracy and convergence}; for example, algorithms constructed based on the regularized problems may offer superior convergence and lower asymptotical errors w.r.t. algorithms built based on the original problem, even though the set of optimal solutions  is explicitly perturbed. This line of thought stemmed in the static domain from the seminal works~\cite{Nesterov2005,Nedic2011,Devolder2011}, and more recently in the online setting~\cite{Simonetto2014d, Bastianello2020asi}.

\begin{figure}
\centering
\includegraphics[width=0.6\textwidth]{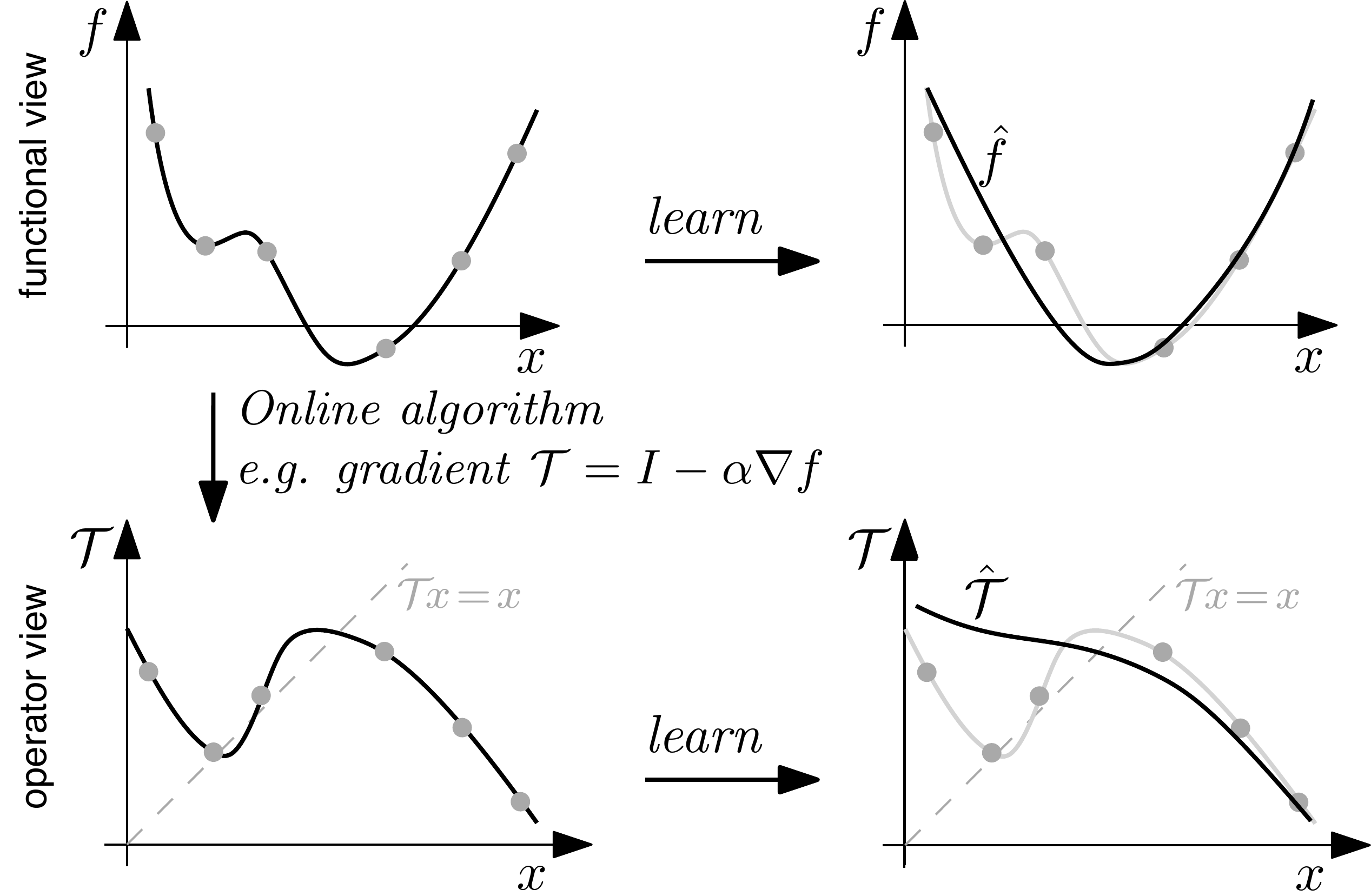}
\caption{The idea of boosting via projection onto the space of ``good'' functions or ``good'' fixed point operators. One can interpret the evaluation of function $f$ or operator $\mathcal{T}$ as noisy evaluations of an underlying ``better'' function or operator, $\hat{f}$ and $\hat{\mathcal{T}}$, respectively, and use the latter to solve the problem instead. This gives rise to convex-regression-based boosting or operation-regression-based boosting (OpReg-Boost).}
\label{fig.1}
\end{figure}

By building on this, a natural question is ``\emph{how to best design a surrogate algorithm that allows a gain in convergence rate without compromising optimality?}''.

To answer this question, one possibility is to modify the cost function by substituting it with a surrogate function that is, for example, strongly convex and smooth.  To fix the idea,  consider a non-convex function $f: \mathbb{R} \rightarrow \mathbb{R}$ as in Figure~\ref{fig.1}. One can evaluate the function at specific points (grey dots) and fit the functional evaluations with a strongly convex function $\hat{f}$. As long as $f$ and $\hat{f}$ are not ``dramatically different'', the reasoning is that solving the problem of minimizing $\hat{f}$ instead of $f$ will then give the  algorithm a boost in terms of convergence rate (without leading to a larger asymptotical error). For this option, which we term Convex Regression, see \cite{arxiv-version}. 

In this paper, we focus on a different approach that consists in  modifying the algorithmic map $\mathcal{A}_k$. The idea is to substitute $\mathcal{A}_k$ with a surrogate mapping that is the ``closest'' to $\mathcal{A}_k$ (in a well defined sense) and has given desirable properties; for example, it is a contractive map. In Figure~\ref{fig.1}, as an example we consider the case of a gradient descent algorithm in terms of a fixed point operator $\mathcal{A}_k = \T_k = \I - \alpha \nabla_{\x} f_k$, with $\alpha>0$ being the step size. The idea here is to use evaluations of $\T_k$ to fit a mapping $\hat{\T}_k$ with useful properties (\emph{e.g.}, contractivity). By using $\hat{\T}_k$ in lieu of $\T_k$, then one may be able to boost convergence and possibly reduce the asymptotical error. We show in our numerical experiments that this  methodology -- referred to as \emph{OpReg-Boost} -- outperforms the first option where one utilizes a surrogate cost. Overall, this paper offers the following contributions.

\begin{enumerate}
\item We present a novel OpReg-Boost method to \emph{learn-project-and-solve} with linear convergence optimization problems. The method is based on operator regression, and it is designed to boost convergence without necessarily increasing the asymptotical error. Operator regression is formulated as a convex quadratically-constrained quadratic programs (QCQPs). 

\item We present efficient ways to solve the operator regression problems in dimension $n$ with $\ell$ observations via a pertinent reformulation of the Peaceman-Rachford splitting (PRS) method, see \emph{e.g.} \cite{bauschke_convex_2017}. Our PRS method is trivially parallel and allows for a reduction of the per-iteration complexity from a convex QCQP in $O(n \ell)$ variables and $O(\ell^2)$ constraints (\emph{i.e.}, a complexity of at least $O((n \ell)^3)$, to $O(\ell^2)$ 1-constraint convex QCQPs in $O(n)$ variables. Importantly, we show that these simpler QCQPs admit a closed form solution, which leads to a per iteration complexity of $O(\ell^2 n)$.

\item We test the performance of the proposed method for two optimization problems: i) a linear regression problem with an ill-conditioned cost [c.f.~\cite{pmlr-v119-mai20a}]; and, ii) an online phase retrieval problem, which requires the minimization of a weakly convex function [c.f.~\cite{duchi_stochastic_2018}]. The proposed operator regression method shows promising performance in both scenarios as compared to forward-backward (with and without backtracking line search) and its accelerated variants FISTA \cite{beck_fast_2009} and an online version of the Anderson acceleration in \cite{pmlr-v119-mai20a}.

\item In the Appendix of \cite{arxiv-version}, we discuss an alternative version of OpReg-Boost that leverages an interpolation technique that goes back to 1945 to trade-off accuracy and speed in operation regression, by interpolating the learned mapping outside the data points via alternating projections. And, also in the Appendix, we present the Convex Regression method, which is also novel (but less performing) and represents an additional contribution in the context of the design of surrogate cost functions.  
\end{enumerate}

The extended version of this manuscript with appendix and proofs can be found in \cite{arxiv-version}.

\subsection{Related work}
Learning to optimize and regularize is a growing research topic; see~\cite{Meinhardt2017,Nghiem2018,Ongie2020,Banert2020,Cohen2020,Pesquet2020,simonetto2019personalized,WotaoL2O} as representative works, even though they focus on slightly different problems. Additional works in the context of learning include the design of convex loss functions in, \emph{e.g.},~
\cite{ramaswamy2016convex,finocchiaro2021unifying}. Interpreting algorithms as mappings and operators (averaged, monotone, \emph{etc.}) has been extremely fruitful for characterizing their convergence properties~\cite{Rockafellar1976, Eckstein1989, bauschke_convex_2017,Ryu2015,Sherson2018}.

Convex regression is treated extensively in~\cite{Seijo2011,Lim2012,Mazumder2019,Blanchet2019}, while recently being generalized to smooth strongly convex functions~\cite{simonetto_smooth_2021} based on A. Taylor's works~\cite{Taylor2016, Taylor2017}; an interesting approach using similar techniques for optimal transport is offered in~\cite{Paty2020}. Operator regression is a recent and at the same time old topic. We are going to build on the recent work~\cite{ryu_operator_2020} and the F.A. Valentine's 1945 paper~\cite{valentine_lipschitz_1945}.

The Anderson acceleration scheme that we compare with is covered in~\cite{pmlr-v119-mai20a} (see also~\cite{Scieur2018,Zhang2020}). 

And finally, the class of weakly convex functions is broad and important in optimization ~\cite{Rockafellar1982,Vial1983,duchi_stochastic_2018,Davis2019,pmlr-v119-mai20b}. Applications featuring this class include robust phase retrieval and many others. {In control theory, this functions extend, e.g., online identification and control to a class on non-linear dynamical systems, and potential games to hypo-monotone settings.}

\section{Learning to accelerate with operator regression}\label{sec:opreg}
Consider the time-varying problem~\eqref{eq:base-problem} and  an associated online algorithm $\mathcal{A}_k$, designed  to track the  optimizers of the problem. The mapping $\mathcal{A}_k$ can be written as sum and/or  composition of maps. To fix the ideas and notation, we provide the following example, which will be used throughout the paper to concretely convey ideas (although we note that the proposed methodology is more widely applicable).

\textbf{Example.} Consider an online forward-backward type algorithm with updates $\x_{k} = \mathcal{A}_k(\x_{k-1})$, where  
\begin{equation}\label{eq:fb}
\mathcal{A}_k = \prox_{\alpha g_k} \circ \T_k , \hspace{.5cm} \T_k := \I - \alpha \nabla_{\x} f_k
\end{equation}
where $ \prox_{\alpha g_k}$ is the proximal operator ($\prox_{\alpha g_k}(\y) = \arg\min_{\x} \{g_k(\x) + \|\x-\y\|^2 / (2\alpha)\}$) and $\I$ the identity map. The properties  of this algorithm depend on the map $\T_k$. In case of a generic smooth non-convex $f_k$ or for convex functions, one can show convergence of the regret to a bounded error~\cite{Simonetto20XX,hallak2020regret}; on the other hand, if $f_k \in \mathcal{S}_{\mu,L}(\R^n)$ uniformly in $k$, $\mu > 0$, then~\eqref{eq:fb} can obtain a linear convergence for the sequence $\{\x_{k}\}$ to the unique optimizer's trajectory of $F_k$ up to a bounded error~\cite{SPM}.
\hfill $\blacklozenge$

Our goal can be formulated as follows: if the algorithmic map $\mathcal{A}_k$ is \emph{not} contractive or is only locally contractive, is it possible to find an approximate mapping $\hat{\mathcal{A}}_k$ that is globally contractive to boost the convergence to the optimal solutions (within an error)?

Consider again the proximal-gradient method in the Example, where we recall that $\mathcal{A}_k = \prox_{\alpha g_k} \circ \mathcal{T}_k$, with $\mathcal{T}_k := \I - \alpha \nabla_{\x} f_k$. When $f_k$ is $\mu$-strongly convex and $L$-smooth uniformly in time, and $\alpha < 2/L$, the mapping $\mathcal{T}_k$ is contractive;  \emph{i.e.}, $\|\T_k(\x) - \T_k(\y)\| \leq \zeta \|\x - \y\|$ for all $\x, \y \in \R^n$ and $\zeta \in (0,1)$. Thus, the recursion $\x_{k} = \mathcal{A}_k (\x_{k-1})$ achieves linear convergence.  However, the question we pose here is the following: when $f_k$ is not $\mu$-strongly convex,  can we still learn map $\hat{\mathcal{T}}_k$, and use the surrogate algorithm $\hat{\mathcal{A}}_k = \prox_{\alpha g_k} \circ \hat{\T_k}$ to achieve linear convergence?\footnote{Notice that the proximal of $g_k$ -- which may encode important properties such as sparsity or constraints -- is not subjected to the learning procedure and remains unchanged.}
 
To this end, using Fact~2.2 in \cite{ryu_operator_2020}, it follows that a mapping $\T_k$ is $\zeta$-contractive interpolable (and therefore extensible to the whole space) if and only if it satisfies:
\begin{equation}
	\norm{\T_k(\x_i)- \T_k(\x_j)}^2 \leq \zeta^2 \norm{\x_i - \x_j}^2, \, \forall i,j \in I_{\ell}, \ i \neq j
\end{equation}
where $I_\ell := \{1, \ldots, \ell\}$ is a finite set of indexes for the points $\{\x_i \in \mathbb{R}^n, i \in I_\ell\}$.  Therefore, using a number of evaluations $\{\T_k(\x_i)\}$ of the mapping $\T_k$ at the points $\{\x_i\}$, we  pose the following convex QCQP as our operator regression problem:
\begin{equation}\label{eq:operator-regression}
\begin{split}
	\hat{\tv} &= \argmin_{ \R^{n\ell} \ni \tv = [\tv_i]_{i \in I_{\ell}}} \frac{1}{2} \sum_{i \in I_{\ell}} \norm{\tv_i - \y_i}^2 \\
	& ~~~~~~~~~~~~ \text{s.t.} \ \norm{\tv_i - \tv_j}^2 \leq \zeta^2 \norm{\x_i - \x_j}^2 \ \forall i,j \in I_{\ell}, i \neq j,
\end{split}
\end{equation}
where the cost function represents a least-square criterion  on the ``observations'' (\emph{i.e.}, the evaluations of the mapping) $\y_i := \T_k(\x_i)$, $i \in I_\ell$, and the constraints enforce contractivity. In particular, the optimal values $\hat{\tv}$ on the data points represent the evaluations of a $\zeta$-contracting operator when applied to those points, \emph{i.e.}, $\hat{\tv}_i = \hat{\T}_k(\x_i)$.

\subsection{PRS-based solver}

The convex problem~\eqref{eq:operator-regression} can be solved using off-the-shelf solvers for convex programs; however, the computational complexity may be a limiting factor, since the problem has a number of constraints that scales quadratically with the number of data points $\ell$. In particular, the computational complexity of interior-point methods would scale at least as $O((n\ell)^3)$. This is generally the case in non-parametric regression~\cite{Mazumder2019,Aybat2014}. To resolve this issue, we propose a parallel algorithm that solves~\eqref{eq:operator-regression} more efficiently based on the so-called Peaceman-Rachford splitting (PRS), see \emph{e.g.} \cite{bauschke_convex_2017}, and that leverages the closed form solution of particular 1-constraint QCQPs.

To this end, define the following set of pairs 
$
	\mathcal{V} = \left\{ e = (i,j) \ | \ i, j \in I_{\ell}, \ i < j \right\}
$
which are ordered (that is, for example we take $(1,2)$ and not $(2,1)$, to avoid counting the pair twice). We associate with each pair $e = (i,j)$ the constraint $\norm{\tv_i - \tv_j}^2 \leq \zeta^2 \norm{\x_i - \x_j}^2$, for a total of $\ell(\ell-1)/2$ constraints. 

Let $\tv_{i,e}$ and $\tv_{j,e}$ be copies of $\tv_i$ and $\tv_j$ associated to the $e$-th pair; then we can equivalently reformulate problem~\eqref{eq:operator-regression} as
\begin{align}\label{eq:equivalent-problem}
	&\min_{\tv_{i,e}, \tv_{j,e}} \frac{1}{2 (\ell-1)} \sum_{e \in \mathcal{V}} \norm{\begin{bmatrix} \tv_{i,e} \\ \tv_{j,e} \end{bmatrix} - \begin{bmatrix} \y_i \\ \y_j \end{bmatrix}}^2 \quad \text{s.t.} \begin{array}{l} \norm{\tv_{i,e} - \tv_{j,e}}^2 \leq \zeta^2 \norm{\x_i - \x_j}^2 \\
	\tv_{i,e} = \tv_{i,e'} \ \forall e, e' | i \sim e, e'\end{array},
\end{align}
where we write that $i \sim e$ if the $e$-th constraint involves $\tv_i$, and recall that $\y_i := \T_k(\x_i)$, $i \in I_\ell$, $i \in I_\ell$. Problem~\eqref{eq:equivalent-problem} is a strongly convex problem with convex constraints defined in the variables $\tv_{i,e}$. Problem~\eqref{eq:equivalent-problem} is in fact a consensus problem which can be decomposed over the pairs $\mathcal{V}$ by using PRS, as defined in the following lemma.

\begin{lemma}\label{lemma.prs}
Problem~\eqref{eq:equivalent-problem} can be solved by using Peaceman-Rachford splitting (PRS), yielding the following iterative procedure. Given the penalty $\rho > 0$, apply for $h \in \N$:
\begin{subequations}\label{eq:prs-solver}
\begin{align}
    \begin{bmatrix} \tv_{i,e}^h \\ \tv_{j,e}^h \end{bmatrix} &= \argmin_{\tv_{i,e},\tv_{j,e}} \left\{ \frac{1}{2 (\ell-1)} \norm{\begin{bmatrix} \tv_{i,e} \\ \tv_{j,e} \end{bmatrix} - \begin{bmatrix} \y_i \\ \y_j \end{bmatrix}}^2 + \frac{1}{2\rho} \norm{\begin{bmatrix} \tv_{i,e} \\ \tv_{j,e} \end{bmatrix} - \z_e^h}^2 \right\} \label{eq:prs-local-update} \\
	&\quad \text{s.t.} \quad \norm{\tv_{i,e} - \tv_{j,e}}^2 \leq \zeta^2 \norm{\x_i - \x_j}^2 \nonumber \\
	\vv_{i,e}^h &= \frac{1}{\ell-1} \sum_{e' | i \sim e'} \left( 2 \tv_{i,e'}^h - \z_{e',i}^h\right), \qquad
	\z_e^{h+1} = \z_e^h + \begin{bmatrix} \vv_{i,e}^h - \tv_{i,e}^h \\ \vv_{j,e}^h - \tv_{j,e}^h \end{bmatrix}.
\end{align}
\end{subequations}
At each iteration, the algorithm solves in parallel $\ell (\ell-1)/2$ convex QCQPs -- each in $2n$ variables and $1$ constraint -- and then aggregates the results. Importantly, the following lemma shows that 1-constraint QCQPs can be solved in closed form with a complexity of $O(n)$, and hence the total per iteration complexity of~\eqref{eq:prs-solver} is $O(\ell^2 n)$.
\end{lemma}

{\bf Proof. } See Appendix~\ref{ap.proof-prs}. \hfill $\blacklozenge$

\begin{lemma}[Solving 1-constraint QCQPs]\label{lem:1-constraint-qcqp}
Consider the (prototypical) QCQP with one constraint
\begin{equation}\label{eq:stylized-local-update}
	(\tv_i^*, \tv_j^*) = \argmin \frac{1}{2} \norm{\begin{bmatrix} \tv_i \\ \tv_j \end{bmatrix} - \begin{bmatrix} \w_i \\ \w_j \end{bmatrix}}^2 \qquad \text{s.t.} \quad \frac{1}{2} \norm{\tv_i - \tv_j}^2 - b \leq 0
\end{equation}
where $b > 0$, which includes as a particular case the update~\eqref{eq:prs-local-update}. Problem~\eqref{eq:stylized-local-update} admits  the following closed form solution
\begin{align}\label{eq:closed-form-opreg}
    \lambda^* &= \max\left\{ 0, \frac{1}{2} \left( \frac{\norm{\w_i - \w_j}}{\sqrt{2 b}} - 1 \right) \right\}, \qquad
	\begin{bmatrix} \tv_i^* \\ \tv_j^* \end{bmatrix} = \frac{1}{1 + 2 \lambda^*}
	\left( \begin{bmatrix}
		1 + \lambda^* & \lambda^* \\ \lambda^* & 1 + \lambda^*
	\end{bmatrix} \otimes \Im_n \right) \begin{bmatrix} \w_i \\ \w_j \end{bmatrix}.
\end{align}
\end{lemma}

{\bf Proof. } See Appendix~\ref{ap.proof-prs}. \hfill $\blacklozenge$

Leveraging Lemma~\ref{lem:1-constraint-qcqp}, we see that~\eqref{eq:prs-local-update} has the following closed form solution
\begin{subequations}
\begin{align}
   & \begin{bmatrix} \w_{i,e}^h \\ \w_{j,e}^h \end{bmatrix} \!=\! \frac{1}{\ell\!-\!1 \!+ \!\rho} \!\left(\! \rho \begin{bmatrix} \y_i \\ \y_j \end{bmatrix} + (\ell\!-\!1) \z_e^h  \!\right)\!,\quad \begin{bmatrix} \tv_{i,e}^h \\ \tv_{j,e}^h \end{bmatrix} \!=\! \frac{1}{1 \!+\! 2 \lambda_e^h}
	\!\left(\! \begin{bmatrix}
		1\! + \!\lambda_e^h & \lambda_e^h \\ \lambda_e^h & 1 + \lambda_e^h
	\end{bmatrix} \!\otimes \!\Im_n \!\right) \!\begin{bmatrix} \w_{i,e}^h \\ \w_{j,e}^h \end{bmatrix},\\
   &\textrm{with}\ \lambda_e^h = \max\left\{ 0, \frac{1}{2} \left( \frac{\norm{\w_{i,e}^h - \w_{j,e}^h}}{\zeta \norm{\x_i - \x_j}} - 1 \right) \right\}.
\end{align}
\end{subequations}

Finally, we discuss how the closed form solution of 1-constraint QCQPs leads to a very low per iteration complexity.

\begin{lemma}[Computational complexity]\label{lem:complexity}
Consider the Peaceman-Rachford splitting~\eqref{eq:prs-solver} that solves the operator regression problem~\eqref{eq:equivalent-problem}, and further notice that the 1-constraint QCQPs~\eqref{eq:prs-local-update} have a closed form solution described in Lemma~\ref{lem:1-constraint-qcqp}.

Then, the computational complexity of the PRS solver is $O(\ell^2 n)$ per iteration. In particular, when the budget of operator calls $\ell$ is much smaller than the dimension of the problem ($n \gg \ell$), then the complexity reduces to $O(n)$ per iteration.
\end{lemma}

{\bf Proof. } See Appendix~\ref{ap.proof-prs}. \hfill $\blacklozenge$

\section{OpReg-Boost}\label{sec:online-opreg}

We are now ready to present our main algorithm. To convey ideas concretely, we focus here on online algorithms of the forward-backward type as in Example~1, \emph{i.e.},\footnote{Access to an operator is the only requirement for the application of OpReg-Boost. However, it is instructive to fix the ideas on a concrete mapping by focusing on the forward-backward algorithm.}
\begin{equation}
\x_{k} = \prox_{\alpha g_k} \left( \T_k(\x_{k-1}) \right), \quad k \in \N,\ \alpha >0.
\end{equation}
where we recall that $\T_k = \I - \alpha \nabla_{\x} f_k$. In particular, we will utilize the operator regression method on the mapping $\T_k$. We recall that the $\prox$ operator is non-expansive; therefore, the Lipschitz constant of the overall mapping $\prox \circ \T_k$ depends on the mathematical properties of $\T_k$ \cite{bauschke_firmly_2012} and, more specifically, of the function $f_k$. In particular, since $f_k$ is not assumed to be strongly convex in general, $\T_k$ may not be contractive and, consequently, $\prox \circ \T_k$ is not contractive either. With this in mind, the goal is to learn a contracting mapping $\hat{\T}_k$ from evaluations of $\T_k$ at some points. The OpReg-Boost algorithm can thus be described as follows. 

\smallskip
\smallskip
\hrule
{\bf OpReg-Boost algorithm}
\smallskip
\hrule \footnotesize
{\bf Required:} number of points $\ell$, stepsize $\alpha$, contraction factor $\zeta$, initial condition $\x_0$.

At each time $k$ do:
\begin{enumerate}
	\item[\textbf{[S1]}] Learn the closest contracting operator to $\T_{k}$, say $\hat{\T}_k$ by:
	\begin{itemize}
		\item[\textbf{[S1.1]}] Choose $\ell-1$ points $\{\x_p\}$ around $\x_{k-1}$ to create the set of points $\{\x_i\} :=\{\x_{k-1} \cup \{\x_p\}\}$, $i \in I_{\ell}$, where the map $\T_k$ is to be evaluated.
		\item[\textbf{[S1.2]}] Evaluate the mapping at the data points: $\y_i = \T_{k}(\x_i)$, $i\in I_{\ell}$, \emph{i.e.}, $\y_i = \x_i - \alpha \nabla_{\x} f_k(\x_i)$.
		\item[\textbf{[S1.3]}] Solve~\eqref{eq:operator-regression} on $\{\x_i, \y_i\}$, $i \in I_{\ell}$ with the PRS-based algorithm.
		\item[\textbf{[S1.4]}] Output $\hat{\tv}_k (=\!\hat{\T}_k(\x_{k-1}))$ from the solution of \textbf{[S1.3]}.
\end{itemize}
	\item[\textbf{[S2]}] Compute $\x_{k} =  \prox_{\alpha g_k} (\hat{\tv}_k) $.
\end{enumerate}
\smallskip
\hrule \normalsize
\vspace{0.25cm}

A couple of remarks are in order. First, the computational complexity of the overall algorithm is dominated by the operation regression problem~\eqref{eq:operator-regression} in step [S1.3]; on the other hand, the number of gradient calls (used to evaluate $\T_k$) is $\ell$-times the one of a standard forward-backward algorithm. At each time $k$, we perform $\ell$ gradient evaluations at the points $\x_{k-1} \cup \{\x_p\}$ (the points $\{\x_p\}$ could be obtained, \emph{e.g.}, by adding a zero-mean Gaussian noise term to $\x_{k-1}$). $\ell$ can be as small as $3$ in practice.

Second, as one can see from steps [S1.2]--[S1.4], the operation regression problem~\eqref{eq:operator-regression} directly provides the evaluation of the regularized operator at the data point $\x_{k-1}$, since we choose $\x_{k-1}$ to define one of the training points.

\section{Numerical results}\label{sec:numerical}
We present a number of experiments to evaluate the performance of the proposed method\footnote{The experiments were implemented in Python and performed on a computer with Intel i7-4790 CPU, $3.60$GHz, and $8$GB of RAM, running Linux. Code and data are available. The implementation is serial (possible due to the manageable size of the regression problems); future work will look at parallel implementations.}. We consider: \emph{(i)} an ill-conditioned online linear regression with a convex cost (but not strongly convex); \emph{(ii)} an online phase retrieval problem that is weakly convex, and which is characterized by a high computational cost per operator evaluation. {The first example is rather well-studied, at least in the well-conditioned region, and it can be used for example to derive control laws under sparsity requirements~\cite{Ohlsson2010,Jovanovic2013}. The second example has important repercussions in adaptive optics, where phase retrieval techniques are used as building blocks to generate control signals~\cite{massioni:hal-01590216,Antonello:15}}

The metric used in the experiments is the tracking error, characterized as the distance from the ground truth signal $\y_k$ of the solution output by the solvers. By $\y_k$ we denote the signal being tracked via linear regression in section~\ref{subsec:linear-regression} or the phase being retrieved in section~\ref{subsec:phase-retrieval}. We choose the tracking error as a proxy for the distance to the optimizer $\x_k^*$ in line with the work of~\cite{duchi_stochastic_2018}, since \emph{(i)} determining $\x_k^*$ is in general hard to do computationally in the problems we are considering and it may not be unique, and \emph{(ii)} the tracking error very naturally provides insights on how the methods perform in estimating the real signals.

\subsection{Online linear regression}\label{subsec:linear-regression}
We consider the following time-varying problem:
\begin{equation}\label{eq:linear-regression}
	\x_k^* \in \argmin_{\x \in \R^n} \frac{1}{2} \norm{\Am \x - \bv_k}^2 + w \norm{\x}_1,
\end{equation}
with $n = 1000$, $w = 1000$, $\Am \in \R^{n \times n}$ such that $\operatorname{rank}(\Am) = n / 2$ and having maximum and minimum (non-zero) eigenvalues $\sqrt{L}$, $\sqrt{\mu}$. The goal is to reconstruct a signal $\y_k$ with sinusoidal components, $1/3$ of them being zero, from the noisy observations $\bv_k = \Am \y_k + \e_k$, and $\e_k \sim \mathcal{N}(\pmb{0}, 10^{-2} \Im)$. Due to $\Am$ being rank deficient, the cost $f_k$ is convex but not strongly so, and we have $\lambda_\mathrm{max}(\nabla_{\x\x} f_k) / \tilde{\lambda}_\mathrm{min}(\nabla_{\x\x} f_k) = L / \mu$, where $\lambda_\mathrm{max}$ and $\tilde{\lambda}_\mathrm{min}$ are the maximum and minimum non-zero eigenvalues of a matrix. The function $f_k$ changes every $\delta = 0.1 s$.

\begin{figure}[!ht]
\centering
\begin{minipage}{.35\textwidth}
	\centering
	\includegraphics[width=\textwidth]{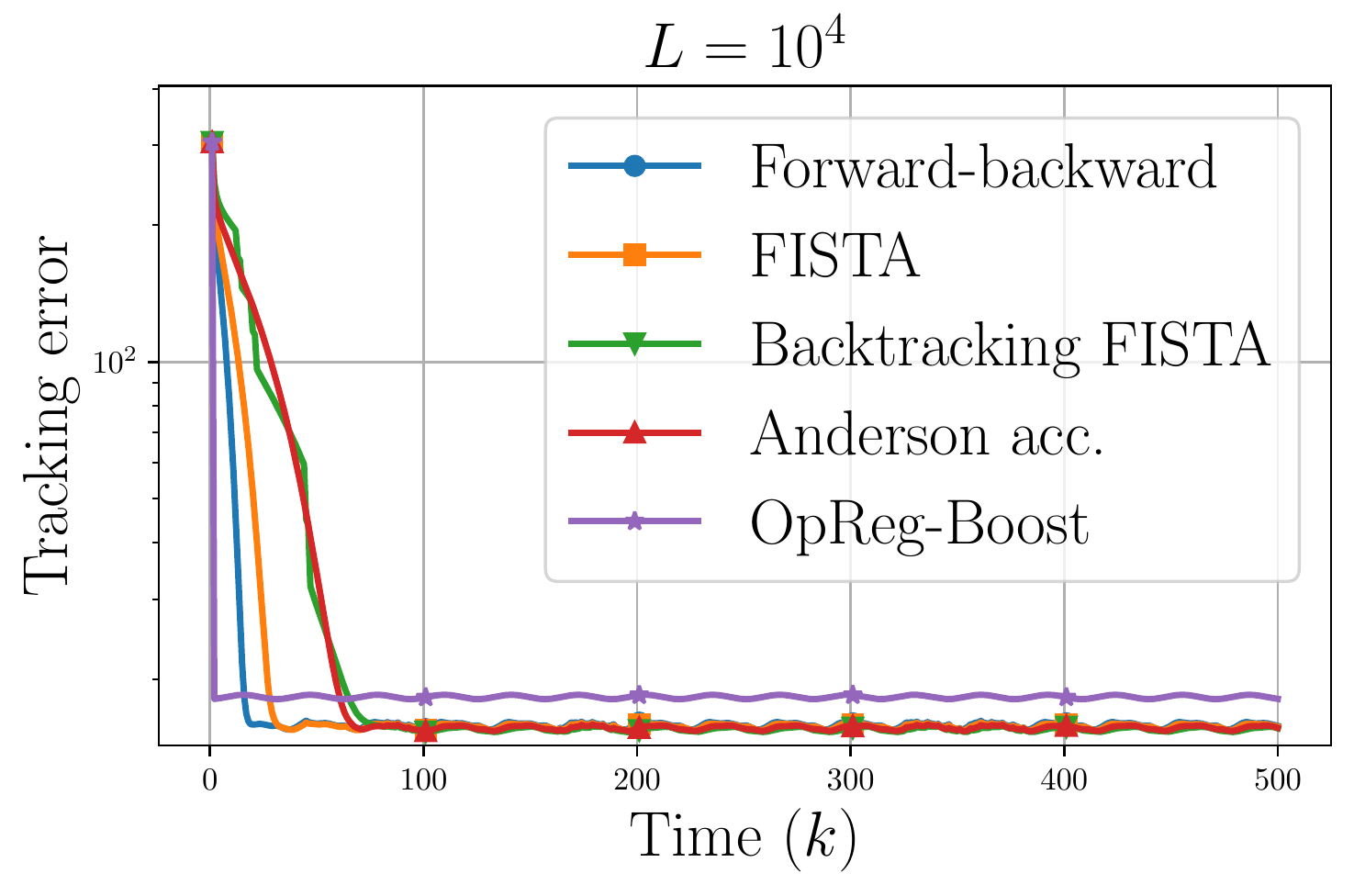}
	\label{fig:lasso-1}
\end{minipage}%
\begin{minipage}{.35\textwidth}
	\centering
	\includegraphics[width=\textwidth]{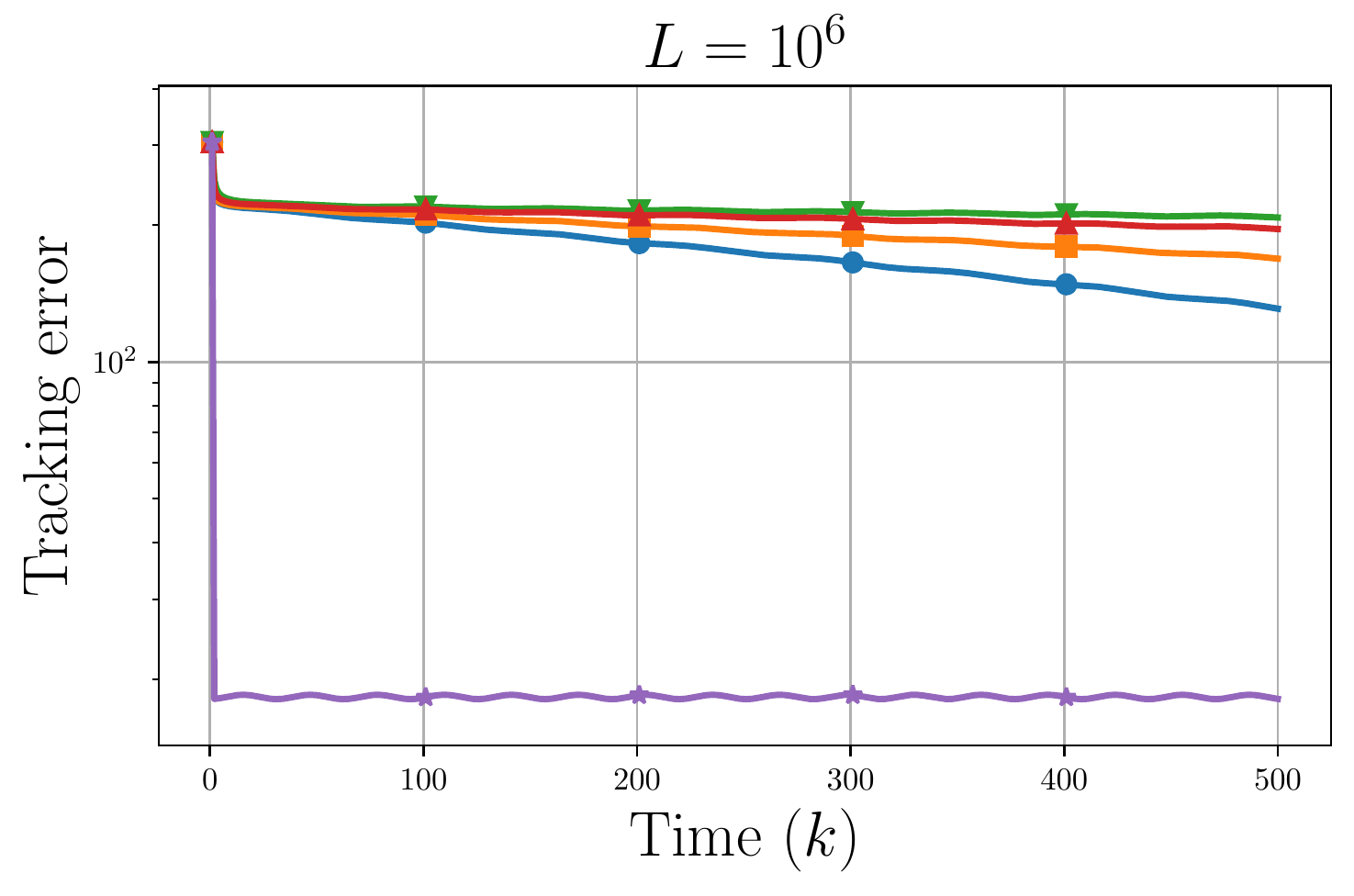}
	\label{fig:lasso-2}
\end{minipage}
\\
\begin{minipage}{.35\textwidth}
	\centering
	\includegraphics[width=\textwidth]{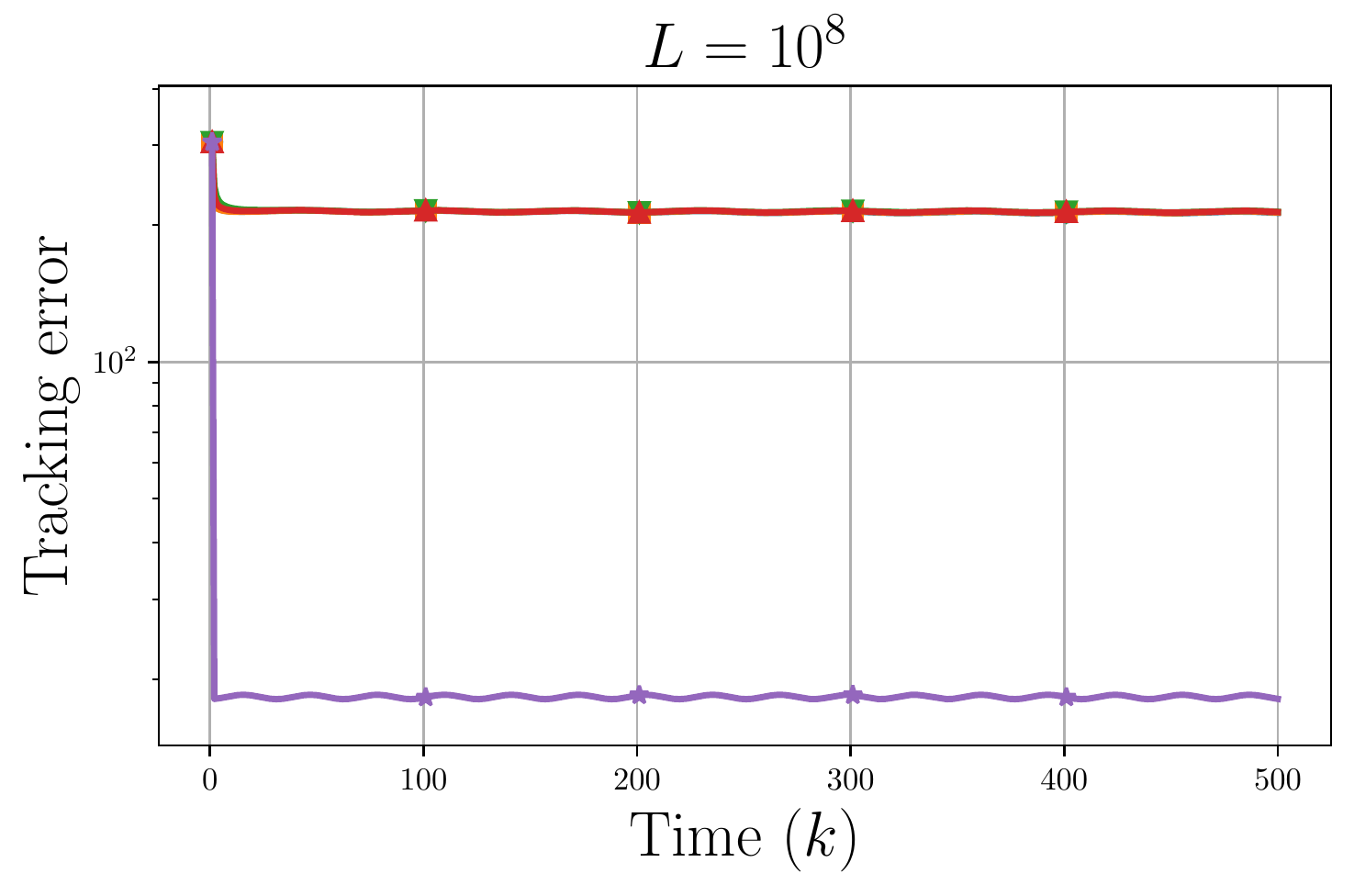}
	\label{fig:lasso-3}
\end{minipage}%
\begin{minipage}{.35\textwidth}
	\centering
	\includegraphics[width=\textwidth]{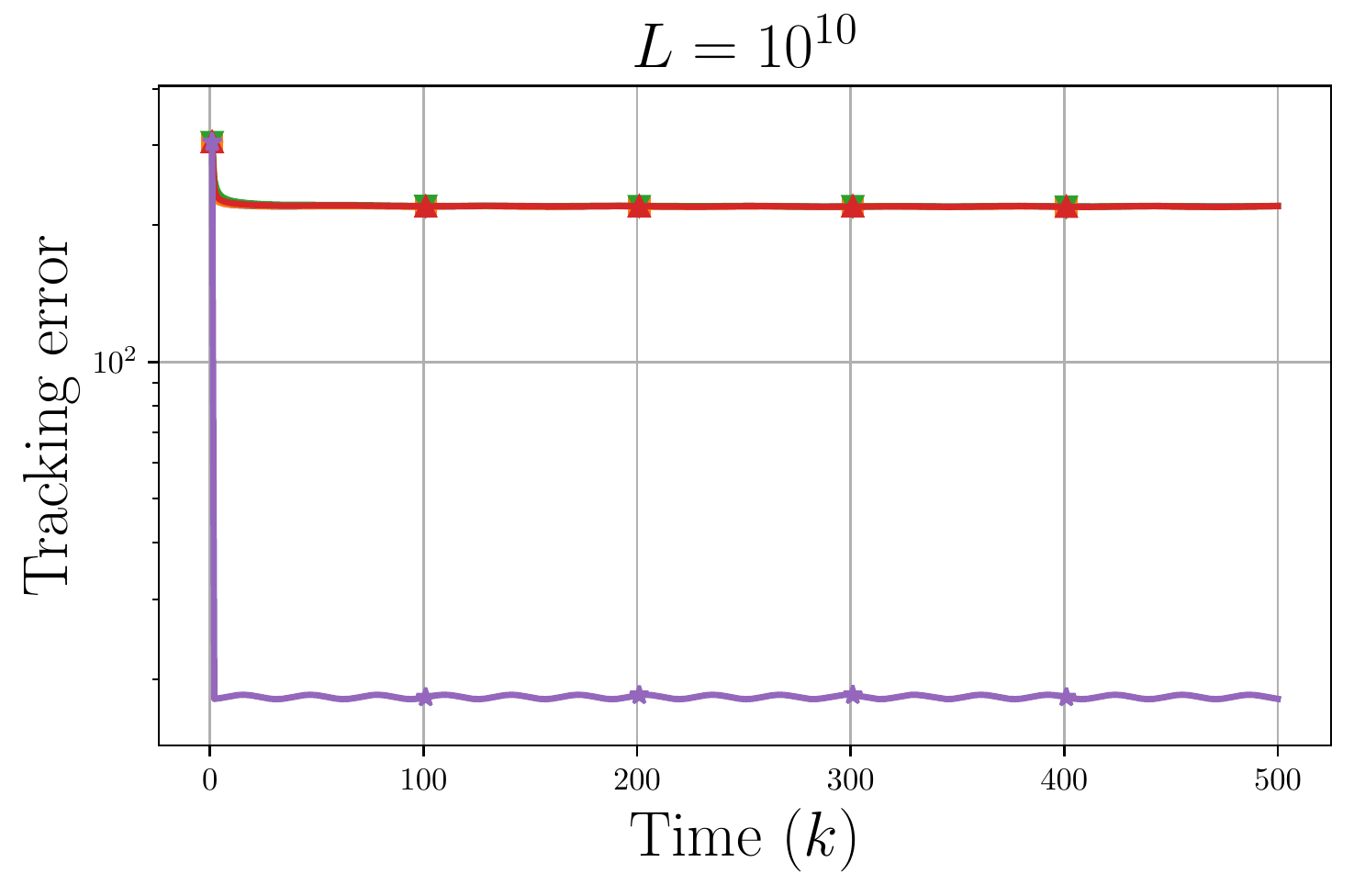}
	\label{fig:lasso-4}
\end{minipage}
\vskip-.5cm
\caption{Comparison with a fixed computational time budget per time $k \in \N$, and for different values of $L$, with fixed $\mu = 1$.}
\label{fig:lasso-condition_num}
\end{figure}

In Figure~\ref{fig:lasso-condition_num}, we show a comparison of the tracking error attained by the proposed OpReg-Boost against the forward-backward method, and its accelerated versions FISTA (with and without backtracking line search) \cite{beck_fast_2009}, and (guarded) Anderson \cite{pmlr-v119-mai20a}. The methods are given the same computational time budget\footnote{Specifically, we evaluate the computational time required by one iteration of OpReg-Boost, and run the other methods for the same time. We remark that OpReg-Boost requires at least the time needed by $\ell$ iterations of forward-backward to generate the operator regression data. For example, our experiments show that with the choice $\rho = 10^{-6}$ during one iteration of OpReg-Boost we can apply $\ell+1$ of forward-backward or FISTA, and one or two of Anderson and FISTA with backtracking.}, the step-size of forward-backward is $\alpha = 2/(L+\mu)$, and the parameters of OpReg-Boost are $\ell = 3$ and $\rho = 10^{-6}$. For large values of $L$ OpReg-Boost outperforms all other methods; in the case $L = 10^4$ it performs slightly worse in terms of asymptotic error, but successfully improves the convergence rate. The reason behind the performance we observe is that as $L$ grows larger, the allowed step-size for forward-backward becomes smaller -- indeed, we have the bound $\alpha < 2 / L$. We further remark that the performance of OpReg-Boost can be improved in the case $L = 10^4$ by choosing a different value of $\rho$, see \cite{arxiv-version}.

{\begin{table}[!ht]
\footnotesize
\vskip-1cm
\caption{Comparison for different values of $n$; for each algorithm we report the asymptotic error (as. err.) and the average computational time \textbf{per step of the algorithm} (t. / s.). We remark that in the simulations \textbf{all methods are given the same computational time budget}, so we apply one or more steps of the algorithm. The simulations are for $L = 10^8$ and $\mu = 1$.}
\label{tab:varying-n}
\centering
\begin{tabular}{lcccccc}
\toprule
& \multicolumn{2}{c}{$n = 10$}  & \multicolumn{2}{c}{$n = 100$} & \multicolumn{2}{c}{$n = 1000$}    \\
\cmidrule(r){2-3} \cmidrule(r){4-5} \cmidrule(r){6-7}
Algorithm & as. err. & t. / s. [s] & as. err. & t. / s. [s] & as. err. & t. / s. [s] \\
\midrule
Forward-backward & $30.00$ & $3.76 \times 10^{-5}$ & $64.88$ & $3.91 \times 10^{-5}$ & $221.36$ & $8.55 \times 10^{-4}$ \\
FISTA & $29.69$ & $3.44 \times 10^{-5}$ & $62.15$ & $4.20 \times 10^{-5}$ & $221.00$ & $8.41 \times 10^{-4}$ \\
FISTA (backtr.) & $29.69$ & $6.33 \times 10^{-4}$ & $62.15$ & $8.27 \times 10^{-4}$ & $220.98$ & $1.77 \times 10^{-2}$ \\
Anderson & $29.69$ & $1.07 \times 10^{-4}$ & $62.16$ & $1.27 \times 10^{-4}$ & $221.01$ & $1.71 \times 10^{-3}$ \\
\textbf{OpReg-Boost} & $\mathbf{2.11}$ & $\mathbf{2.48 \times 10^{-4}}$ & $\mathbf{6.14}$ & $\mathbf{2.98 \times 10^{-4}}$ & $\mathbf{18.72}$ & $\mathbf{2.88 \times 10^{-3}}$ \\
\bottomrule
\end{tabular}
\end{table}}

Finally, in Table~\ref{tab:varying-n} we report the asymptotic error and computational time of OpReg-Boost as compared to the forward-backward based solvers for three different sizes of the problem with $L = 10^8$ and $\mu = 1$. In terms of asymptotic error -- evaluated when all methods are given the same total computational time -- the performance of OpReg-Boost is consistently better than the other methods. Regarding the computational time per step of the algorithm we see that OpReg-Boost is comparable with the accelerated methods FISTA with backtracking and Anderson. On the other hand, the computationally lighter forward-backward and FISTA require less time per step, but, again, when given the same computational time the performance of OpReg-Boost is still better.

\subsection{Online phase retrieval}\label{subsec:phase-retrieval}
We consider now the following  phase retrieval problem presented in \cite{duchi_stochastic_2018}:
\vskip -0.2in
\begin{equation}\label{eq:phase-retrieval}
	\x_k^* \in \argmin_{\x \in \R^n} \frac{1}{m} \sum_{i = 1}^{m} \left| \langle \av_i, \x \rangle^2 - b_{i,k} \right|,
\end{equation}
where the goal is to reconstruct the time-varying signal $\y_k \in \mathbb{S}^{n-1}$, $n = 50$, from the noisy measurements $b_{i,k} = \langle \av_i, \y_k \rangle + \xi_{i,k}$, $i = 1, \ldots, m$ and $m = 100$. The signal $\y_k$ is piece-wise constant, with the value of each constant piece being independently drawn. The additive noises are i.i.d. Laplace with zero mean and scale parameter $1$. The $\av_i$ are the rows of $\Am \in \R^{m \times n}$, constructed as $\Am = \mathbold{U} \mathbold{D}$, with $\mathbold{U} \in \R^{m \times n}$ an orthogonal matrix, and $\mathbold{D} \in \R^{n \times n}$ a diagonal one with elements $L = 10^2$, $\mu = 1$ (hence the condition number of $\Am$ is $10^2$), and the remaining $n - 2$ drawn from $\mathcal{U}[\mu, L]$. The problem changes every $\delta = 1 s$.

We consider the \emph{prox-linear} solver proposed in the work of \cite{drusvyatskiy_error_2018} (see also \cite{duchi_stochastic_2018}), characterized by $\T_k(\y) = \prox_{\alpha f_{k,\y}}(\y)$, where $f_{k,\y}$ denotes the following linearized version of the cost in~\eqref{eq:phase-retrieval}:
$
	f_{k,\y}(\x) = \frac{1}{m} \sum_{i = 1}^m \left| \langle \av_i, \y \rangle^2 + 2 \langle \av_i, \y \rangle \langle \av_i, \x - \y \rangle - b_{i,k} \right|.
$
We choose the step-size of the prox-linear solver as $\alpha = 10^{-3}$, which empirically led to convergence (at least in the initial transient) without the need for line search. Notice that the proximal operator $\T_k$ does not have a closed form, and each operator call requires the solution of a quadratic program, which takes $0.177 \pm 0.0052 s$.

\begin{figure}[!ht]
\centering
\begin{minipage}{.35\textwidth}
	\centering
	\includegraphics[width=\textwidth]{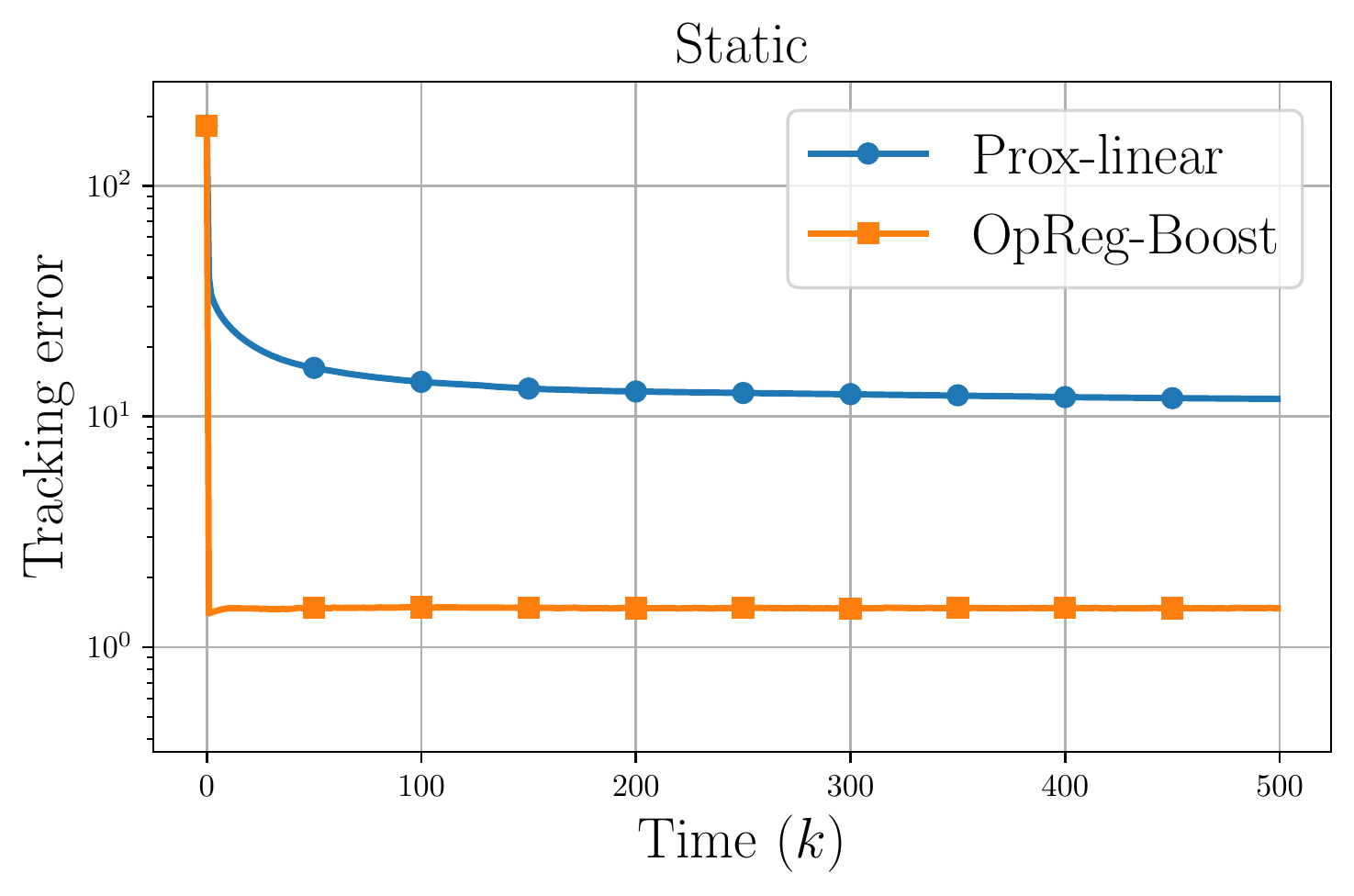}
	\label{fig:phase-1}
\end{minipage}%
\begin{minipage}{.35\textwidth}
	\centering
	\includegraphics[width=\textwidth]{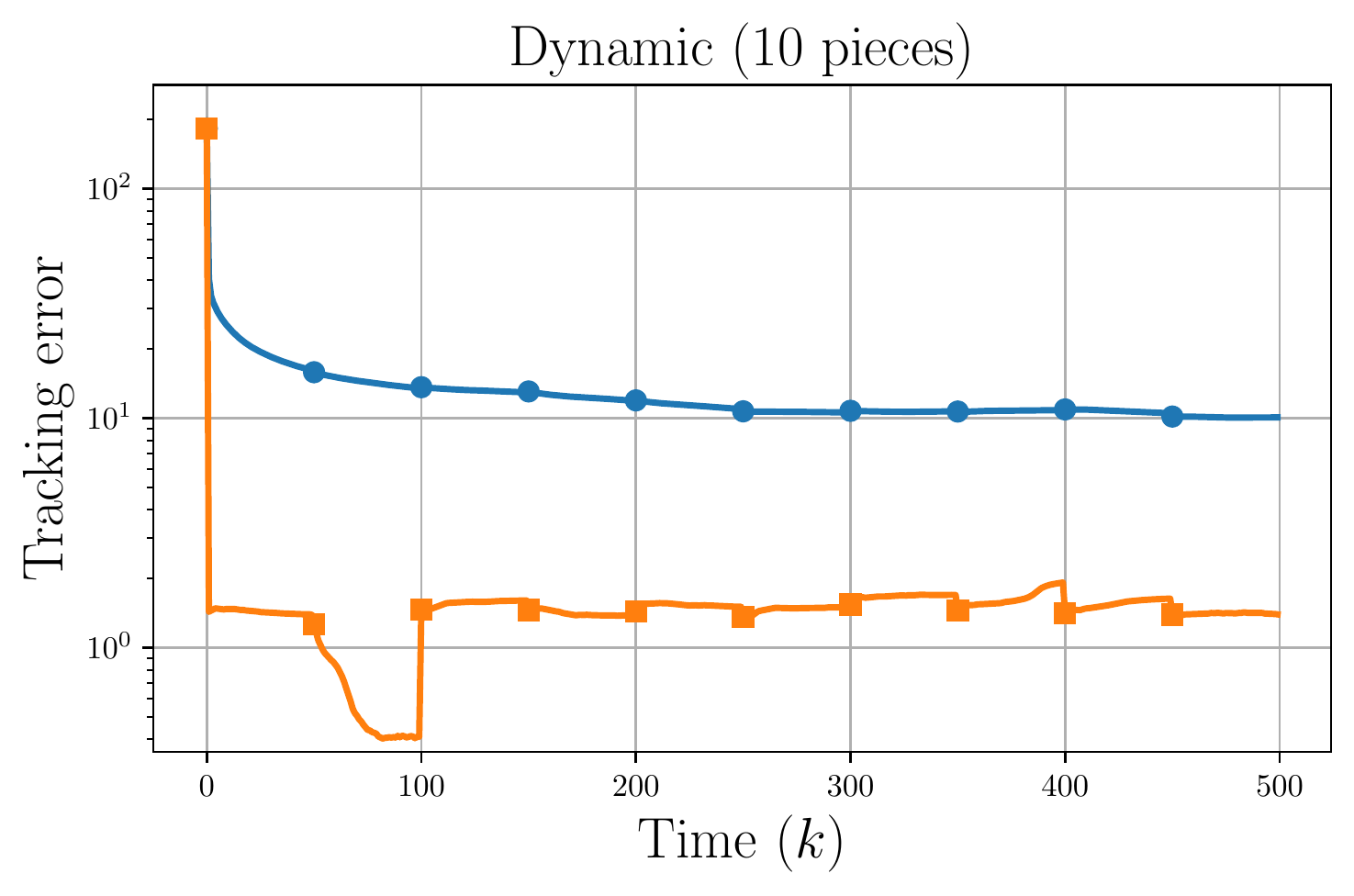}
	\label{fig:phase-2}
\end{minipage}
\\
\begin{minipage}{.35\textwidth}
	\centering
	\includegraphics[width=\textwidth]{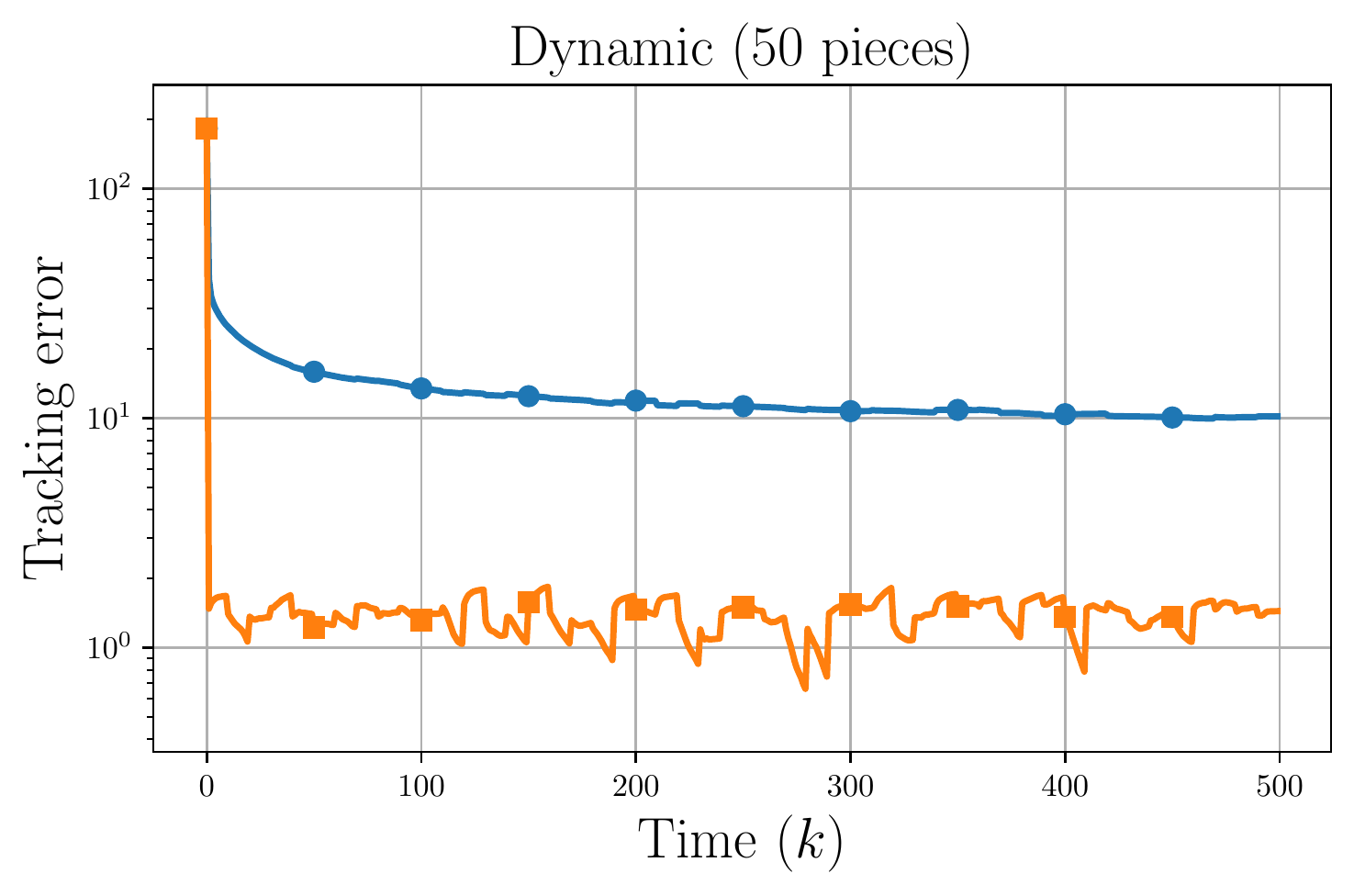}
	\label{fig:phase-3}
\end{minipage}%
\begin{minipage}{.35\textwidth}
	\centering
	\includegraphics[width=\textwidth]{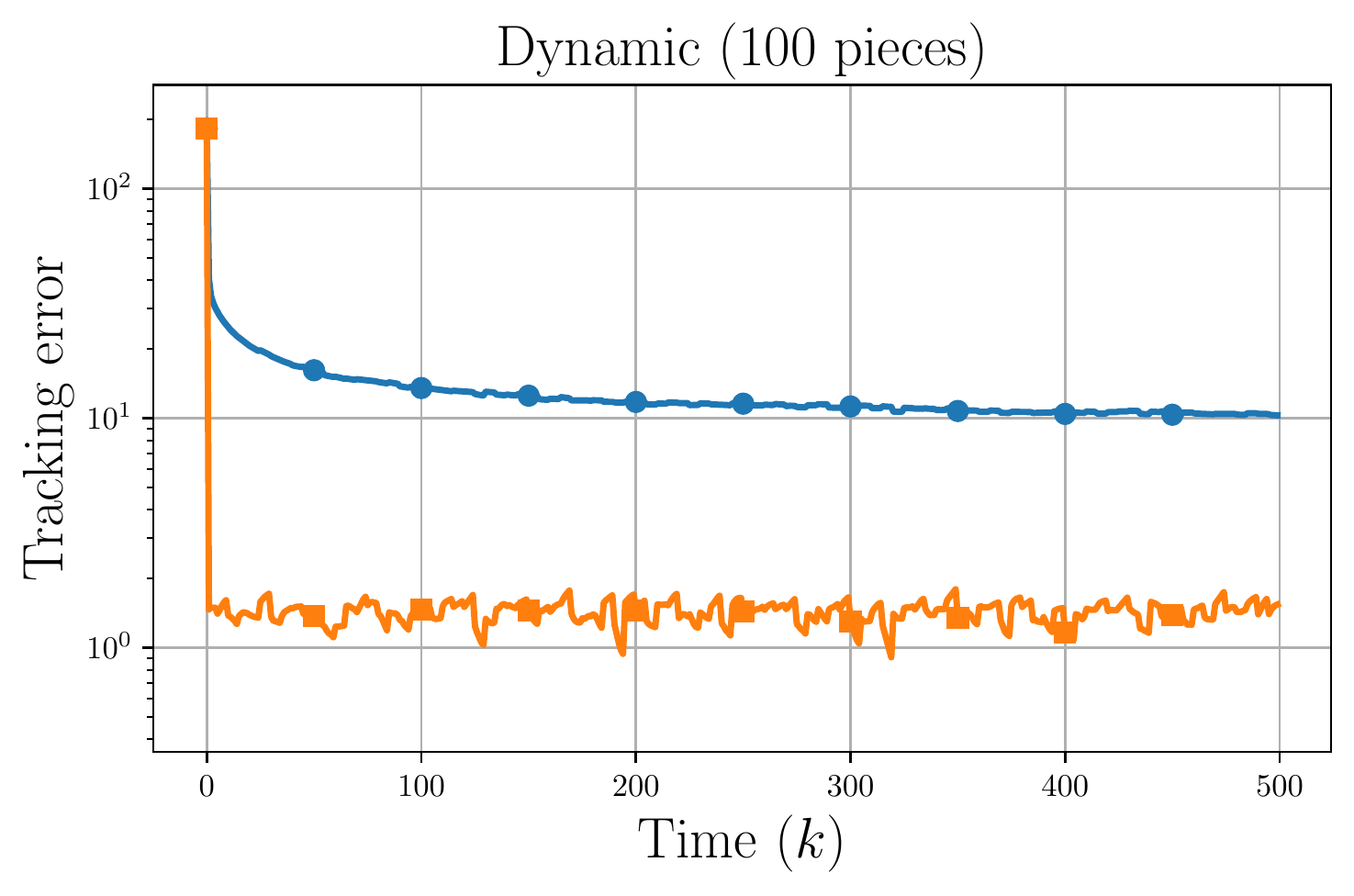}
	\label{fig:phase-4}
\end{minipage}
\caption{Comparison of the tracking error evolution for prox-linear \cite{drusvyatskiy_error_2018} and OpReg-Boost. The methods are tasked with retrieving phase signals that are piece-wise continuous, with different number of pieces in each.}
\label{fig:phase-retrieval}
\end{figure}

We also consider our OpReg-Boost algorithm applied to the operator\footnote{This shows better performance in practice rather than regularizing $\T_k$ alone. Strictly speaking, with this choice, function $g_k$ in~\eqref{eq:base-problem} would be the indicator function of a non-convex set. The good performance of the proposed approach however suggests that it can be applied to more general problems than~\eqref{eq:base-problem}.} $\proj_{\mathbb{S}^{n-1}} \circ \T_k$, and which regularizes the operator $\T_k$ to a $\zeta$-contractive operator, yielding then $\proj_{\mathbb{S}^{n-1}} \circ \hat{\T}_k$. The solution of an operator regression problem requires $1.35 \times 10^{-3} \pm 0.011 s$. In the results, then, during the time before the arrival of a new problem ($\delta = 1 s$), we perform either $4$ steps of the prox-linear solver, or one step of OpReg-Boost with $3$ training points (and choosing the PRS parameter $\rho = 10^{-4}$).

In Figure~\ref{fig:phase-retrieval}, we show the tracking error of prox-linear compared with OpReg-Boost when the signal has different numbers of constant pieces (from being static -- one constant value -- to being highly dynamic -- changing every $5 s$). As we can see, OpReg-Boost consistently outperforms prox-linear, including in the static case, in which OpReg-Boost quickly converges to the (approximate) fixed point, while prox-linear converges more slowly.

\bibliographystyle{IEEEtran}
\bibliography{PaperCollection00,references}

\newpage
\appendix


\section{Convex regression} 

The concept of convex regression is briefly introduced here; we refer the reader to, \emph{e.g.},~\cite{Mazumder2019,simonetto_smooth_2021} for the technical details. Suppose one has $\ell$ noisy measurements of a convex function $\varphi(\x): \R^n \to \R$ (say $y_i$) at points $\x_i \in \R^n$, $i \in I_{\ell}$, and (optionally) its gradients $\nabla_{\x} \varphi(\x_i)$. Then convex regression is a least-squares approach to estimate the generating function based on the measurements. Formally, letting $\varphi \in \mathcal{S}_{\mu,L}(\R^n)$, then one would like to solve the functional estimation problem
\begin{equation}\label{eq.inf}
    \hat{\varphi}_{\ell} \in \argmin_{\psi \in \mathcal{S}_{\mu, L}(\R^n)}\Big\{ \sum_{i\in I_{\ell}} (y_i - \psi(\x_i))^2 + \|\z_i - \nabla_{\x} \psi(\x_i)\|^2 \Big\} \,,
\end{equation}
where $y_i$ and $\z_i$ are the measurements of the function $\varphi$ and its gradient at the data points. 

In lieu of the infinite-dimensional problem~\eqref{eq.inf}, one can consider an equivalent estimation problem to find the true values of the function and its gradients at the data point (\emph{i.e.}, $\f = [\varphi_i]_{i \in I_{\ell}}$, $\bdelta = [\nabla \varphi_i]_{i \in I_{\ell}}$); this amounts to the following convex quadratically-constrained quadratic program:
\begin{subequations}\label{socp}
\begin{eqnarray}
(\f^*, \bdelta^*) &&\hskip-.5cm= \argmin_{\f\in\R^\ell\!,~ \bdelta \in\R^{n\ell}} \frac{1}{2} \sum_{i\in I_{\ell}} (y_i - \varphi_i)^2 + \|\z_i - \bdelta_i \|^2 \\
\mathrm{s.t.:} &&
\varphi_i - \varphi_j - \bdelta_j^\transp (\x_i - \x_j) \geq \frac{1}{2(1 - \mu/L)} \times \\ && \,\, \left(\frac{1}{L}\|\bdelta_i - \bdelta_j \|^2_2 + \mu \|\x_i - \x_j\|_2^2 - 2 \frac{\mu}{L} (\bdelta_j-\bdelta_i)^\transp (\x_j - \x_i) \right), \quad \forall i, j \in I_{\ell}. \nonumber
\end{eqnarray} 
\end{subequations}
Using the point estimate $(\f^*, \bdelta^*)$, an interpolation scheme then extends the estimation of the function to the whole space (maintaining the functional properties) as:  
\begin{equation}\label{interp}
    \hat{\varphi}_{\ell}(\x) = \mathrm{conv}(p_i(\x)) + \frac{\mu}{2} \|\x\|^2_2 \in \mathcal{S}_{\mu,L}(\R^n),
\end{equation}
where  $\mathrm{conv}(\cdot)$ indicates the convex hull and
\begin{equation}
    p_i(\x) := \frac{L-\mu}{2} \| {\x} - \x_i\|_2^2 + (\bdelta_i^*-\mu \x_i)^\transp \x - \bdelta_i^{*,\transp} \x_i + f_i^* + \frac{\mu}{2} \|\x_i\|_2^2 . 
\end{equation}

For the goal of boosting the convergence of online algorithms we can leverage convex regression as follows. For each time $k$ we perform $\ell$ evaluations of the function $f_k(\x)$ (and optionally its gradients) and utilize the procedure above to project the function $f_k(\cdot)$ onto the space of strongly convex and smooth functions $\mathcal{S}_{\mu,L}(\R^n)$. Then, we can use the estimated function $\hat{f}_k \in \mathcal{S}_{\mu,L}(\R^n)$ to build an algorithm $\hat{\mathcal{A}}_k$. For example, for the proximal-gradient method in Example~1, the algorithmic map amounts to $\hat{\mathcal{A}}_k = \prox_{\alpha g_k} \circ (\I - \alpha \nabla_{\x} \hat{f}_k(\cdot))$.

However, the application of convex regression within the tight time constraints of online optimization hinges on the efficient solution of~\eqref{socp}. In the following section we describe a solver based on Peaceman-Rachford splitting that solves~\eqref{socp}.

\subsection{PRS-based solver}
Similarly to operator regression in section~\ref{sec:opreg}, the idea is to make multiple copies of the unknowns $(\varphi_i, \bdelta_i)$ so that problem~\eqref{socp} can be rewritten as a set of simpler QCQPs in $2 (n+1)$ unknowns and with two constraints each.

Recalling the definition of the pairs $\mathcal{V} = \left\{ e = (i,j) \ | \ i, j \in I_{\ell}, \ i < j \right\}$, we can associate with each $e = (i,j)$ the two constraints
\begin{equation}\label{eq:cvx-constraints}
\begin{split}
    \varphi_{i,e} - \varphi_{j,e} - \bdelta_{j,e}^\transp (\x_i - \x_j) &\geq \frac{1}{2(1 - \mu/L)} \left(\frac{1}{L} \|\bdelta_{i,e} - \bdelta_{j,e} \|^2_2 + \mu \|\x_i - \x_j\|_2^2 \right. + \\ & \left. - 2 \frac{\mu}{L} (\bdelta_{j,e}-\bdelta_{i,e})^\transp (\x_j - \x_i) \right) \\
    \varphi_{j,e} - \varphi_{i,e} - \bdelta_{i,e}^\transp (\x_j - \x_i) &\geq \frac{1}{2(1 - \mu/L)} \left(\frac{1}{L} \|\bdelta_{i,e} - \bdelta_{j,e} \|^2_2 + \mu \|\x_i - \x_j\|_2^2 \right. + \\ & \left. - 2 \frac{\mu}{L} (\bdelta_{i,e}-\bdelta_{j,e})^\transp (\x_j - \x_i) \right)
\end{split}
\end{equation}
where by \emph{e.g.} $\varphi_{i,e}$ we denote the copy of $\varphi_i$ in the pair $e = (i,j)$.
Problem~\eqref{socp} then becomes
\begin{equation}\label{eq:cvxreg-equivalent-problem}
\begin{split}
& \min_{\varphi_{i,e} \in \R,~\bdelta_{i,e} \in \R^n} \frac{1}{2 (\ell-1)} \sum_{e \in \mathcal{V}} \norm{\begin{bmatrix} \varphi_{i,e} \\ \varphi_{j,e} \end{bmatrix} - \begin{bmatrix} y_i \\ y_j \end{bmatrix}}^2 + \norm{\begin{bmatrix} \bdelta_{i,e} \\ \bdelta_{j,e} \end{bmatrix} - \begin{bmatrix} \z_i \\ \z_j \end{bmatrix}}^2 \\
&\mathrm{s.t.:} \ \ \eqref{eq:cvx-constraints}, \quad \forall i, j \in I_{\ell} \\
& \qquad (\varphi_{i,e}, \bdelta_{i,e}) = (\varphi_{i,e'}, \bdelta_{i,e'}) \ \forall e, e' | i \sim e, e'.
\end{split}
\end{equation}
We can now leverage the separable structure of problem~\eqref{eq:cvxreg-equivalent-problem} in order to solve it with PRS.

\begin{lemma}
Problem~\eqref{eq:cvxreg-equivalent-problem} can be solved by using Peaceman-Rachford splitting (PRS), yielding the following iterative procedure. Given the penalty $\rho > 0$, apply for $h \in \N$:
\begin{subequations}\label{eq:prs-solver-cvxreg}
\begin{align}
    \tv_e^h &= \argmin_{\tv_e \in \R^{2(n+1)}} \left\{ \frac{1}{2 (\ell-1)} \norm{\tv_e - \begin{bmatrix} \varphi_i \\ \varphi_j \\ \bdelta_i \\ \bdelta_j \end{bmatrix}}^2 + \frac{1}{2\rho} \norm{\tv_e - \z_e^h}^2 \right\} \label{eq:prs-local-update-cvxreg} \\
	&\quad \text{s.t.} \quad \eqref{eq:cvx-constraints} \nonumber \\
	\vv_{i,e}^h &= \frac{1}{\ell-1} \sum_{e' | i \sim e'} \left( 2 \tv_{i,e'}^h - \z_{e',i}^h\right), \qquad
	\z_e^{h+1} = \z_e^h + \begin{bmatrix} \vv_{i,e}^h - \tv_{i,e}^h \\ \vv_{j,e}^h - \tv_{j,e}^h \end{bmatrix}
\end{align}
\end{subequations}
where $\tv_e = [\varphi_{i,e}^h, \varphi_{j,e}^h, \bdelta_{i,e}^h, \bdelta_{j,e}^h]^\top$.
At each iteration, the algorithm solves in parallel $\ell (\ell-1)/2$ convex QCQPs -- each in $2(n+1)$ variables and $2$ constraints -- and then aggregates the results. Importantly, the following lemma shows that the particular 2-constraint QCQPs can be solved in closed form with a complexity of $O(n)$, and hence the total per iteration complexity of~\eqref{eq:prs-solver-cvxreg} is $O(\ell^2 n)$.
\end{lemma}

{\bf Proof. } Follows the same derivation of Lemma~\ref{lemma.prs}. \hfill $\blacklozenge$

\begin{lemma}[Solving~\eqref{eq:prs-local-update-cvxreg}]\label{lem:2-constraint-qcqp}
The update~\eqref{eq:prs-local-update-cvxreg} can be rewritten as the following QCQP with two constraints:
\begin{equation}\label{eq:stylized-local-update-cvxreg}
\begin{split}
	\tv^* &= \argmin \frac{1}{2} \norm{\tv - \w}^2 \\
	&\text{s.t.} \quad \frac{1}{2} \tv^\top \Pm \tv + \langle \tv, \q_1 \rangle + r \leq 0, \quad \frac{1}{2} \tv^\top \Pm \tv + \langle \tv, \q_2 \rangle + r \leq 0
\end{split}
\end{equation}
where $\Pm = \operatorname{blk\,diag}\left\{ 0_{2 \times 2}, \begin{bmatrix} 1 & -1 \\ -1 & 1 \end{bmatrix} \otimes \Im_n \right\}$, $r = L \mu \norm{\x_i - \x_j}^2 / 2$, and
$$
    \w = \frac{1}{\ell-1 + \rho} \left( \rho \begin{bmatrix} \varphi_i \\ \varphi_j \\ \bdelta_i \\ \bdelta_j \end{bmatrix} + (\ell-1) \z_e^h \right), \quad \q_1 = \begin{bmatrix} -(L-\mu) \\ (L-\mu) \\ -\mu (\x_i - \x_j) \\ L (\x_i - \x_j) \end{bmatrix}, \quad \q_2 = \begin{bmatrix} (L-\mu) \\ -(L-\mu) \\ -L (\x_i - \x_j) \\ \mu (\x_i - \x_j) \end{bmatrix}.
$$
The problem~\eqref{eq:stylized-local-update-cvxreg} admits  the following closed form solution
\begin{align}\label{eq:closed-form-opreg-cvxreg}
    \lambda_1^* &= \max\left\{ 0, \frac{1}{2} (\lambda_+^* + \lambda_-^*) \right\}, \quad \lambda_2^* = \max\left\{ 0, \frac{1}{2} (\lambda_+^* - \lambda_-^*) \right\} \\
	\tv^* &= \operatorname{blk\,diag}\left\{ \Im_2, \frac{1}{1 + 2(\lambda_1^* + \lambda_2^*)} \begin{bmatrix} 1+\lambda_1^*+\lambda_2^* & \lambda_1^*+\lambda_2^* \\ \lambda_1^*+\lambda_2^* & 1+\lambda_1^*+\lambda_2^* \end{bmatrix} \otimes \Im_n \right\}  \left( \w - \lambda_1^* \q_1 - \lambda_2^* \q_2 \right)
\end{align}
where
\begin{align}
    \lambda_-^* &= \frac{\langle \x_i - \x_j, [\w]_3 + [\w]_4 \rangle - 2 ([\w]_1 - [\w]_2)}{(L-\mu) (\norm{\x_i - \x_j}^2 + 4)} \\
    \lambda_+^* &= \frac{1}{2} \left( \frac{\norm{(L+\mu) (\x_i - \x_j) - 2 ([\w]_3 - [\w]_4)}}{(L-\mu) \norm{\x_i - \x_j}} \right)
\end{align}
with $[\w]_l$ denoting the $l$-th block in the vector $\w$\footnote{Notice that the first two blocks are scalars, while the third and fourth are vectors in $\R^n$.}.
\end{lemma}

{\bf Sketch of proof. } From the KKT conditions of~\eqref{eq:stylized-local-update-cvxreg} we can easily derive the expression for $\tv^*$, which means that we need to find the optimal solution for the two Lagrange multipliers $\lambda_1$ and $\lambda_2$.

\noindent Assume now that $\lambda_1, \lambda_2 > 0$, then the constraints of~\eqref{eq:stylized-local-update-cvxreg} should be verified as equalities. Substituting the expression for $\tv^*$ into these two equalities and subtracting them we get the expression for $\lambda_-^* = \lambda_1^* - \lambda_2^*$. Summing the two equalities instead yields a quadratic equation in $\lambda_+^* = \lambda_1^* + \lambda_2^*$, whose positive solution yields the expression above.

\noindent Finally, by definition $\lambda_1^* = (\lambda_+^* + \lambda_-^*) / 2$ and $\lambda_2^* = (\lambda_+^* - \lambda_-^*) / 2$, but since $\lambda_+^*$ and $\lambda_-^*$ may be negative quantities, we impose that the resulting Lagrange multipliers be non-negative. \hfill $\blacklozenge$

\begin{lemma}[Computational complexity]\label{lem:complexity-cvxreg}
Consider the Peaceman-Rachford splitting~\eqref{eq:prs-solver-cvxreg} that solves the operator regression problem~\eqref{eq:cvxreg-equivalent-problem}, and further notice that the 2-constraint QCQPs~\eqref{eq:prs-local-update-cvxreg} have a closed form solution described in Lemma~\ref{lem:2-constraint-qcqp}.

Then, the computational complexity of the PRS solver is $O(\ell^2 n)$ per iteration. In particular, when the budget of operator calls $\ell$ is much smaller than the dimension of the problem ($n \gg \ell$), then the complexity reduces to $O(n)$ per iteration.
\end{lemma}

{\bf Proof. } The closed form solution described in Lemma~\ref{lem:2-constraint-qcqp} only requires operation on $2(n+1)$ vectors, with complexity $O(n)$. Since we have $\ell (\ell-1)/2$ such closed forms to compute, then the thesis follows. \hfill $\blacklozenge$

\subsection{CvxReg-Boost}\label{ap.cvx-reg}

We present here the CvxReg-Boost algorithm. 

\smallskip
\smallskip
\hrule
{\bf CvxReg-Boost algorithm}
\smallskip
\hrule
{\bf Required:} number of points $\ell$, stepsize $\alpha$, functional parameters  $\mu, L$, initial condition $\x_0$.

At each time $k$ do:
\begin{enumerate}
	\item[\textbf{[S1]}] Learn the closest function in $\mathcal{S}_{\mu,L}(\R^n)$ to $f_k(\x)$, say $\hat{f}_k$ by:
	\begin{itemize}
		\item[\textbf{[S1.1]}] Choose $\ell-1$ points $\{\x_p\}$ around $\x_{k-1}$ to create the set of points $\{\x_i\} :=\{\x_{k-1} \cup \{\x_p\}\}$, $i \in I_{\ell}$, where the function $f_k(\x)$ is to be evaluated.
		\item[\textbf{[S1.2]}]  Evaluate the function (and optionally its gradients) on the data points: $\varphi_i = f_k(\x_i)$, $i\in I_{\ell}$, $\z_i =  \nabla_{\x} f_k(\x_i)$.
		\item[\textbf{[S1.3]}] Solve~\eqref{socp} using the PRS-based solver~\eqref{eq:prs-solver-cvxreg} and output $\hat{\z}_k (= \nabla_{\x}\hat{f}_k (\x_{k-1}))$.
    \end{itemize}
	\item[\textbf{[S2]}] Apply $\x_{k} = \prox_{\alpha g_k} (\x_{k-1} - \alpha \nabla_{\x}\hat{f}_k (\x_{k-1})) = \prox_{\alpha g_k} (\x_{k-1} - \alpha \hat{\z}_k) $.
\end{enumerate}
\smallskip
\hrule

\subsection{Numerical results}
\begin{figure}[t]
\centering
\includegraphics[width=0.75\linewidth]{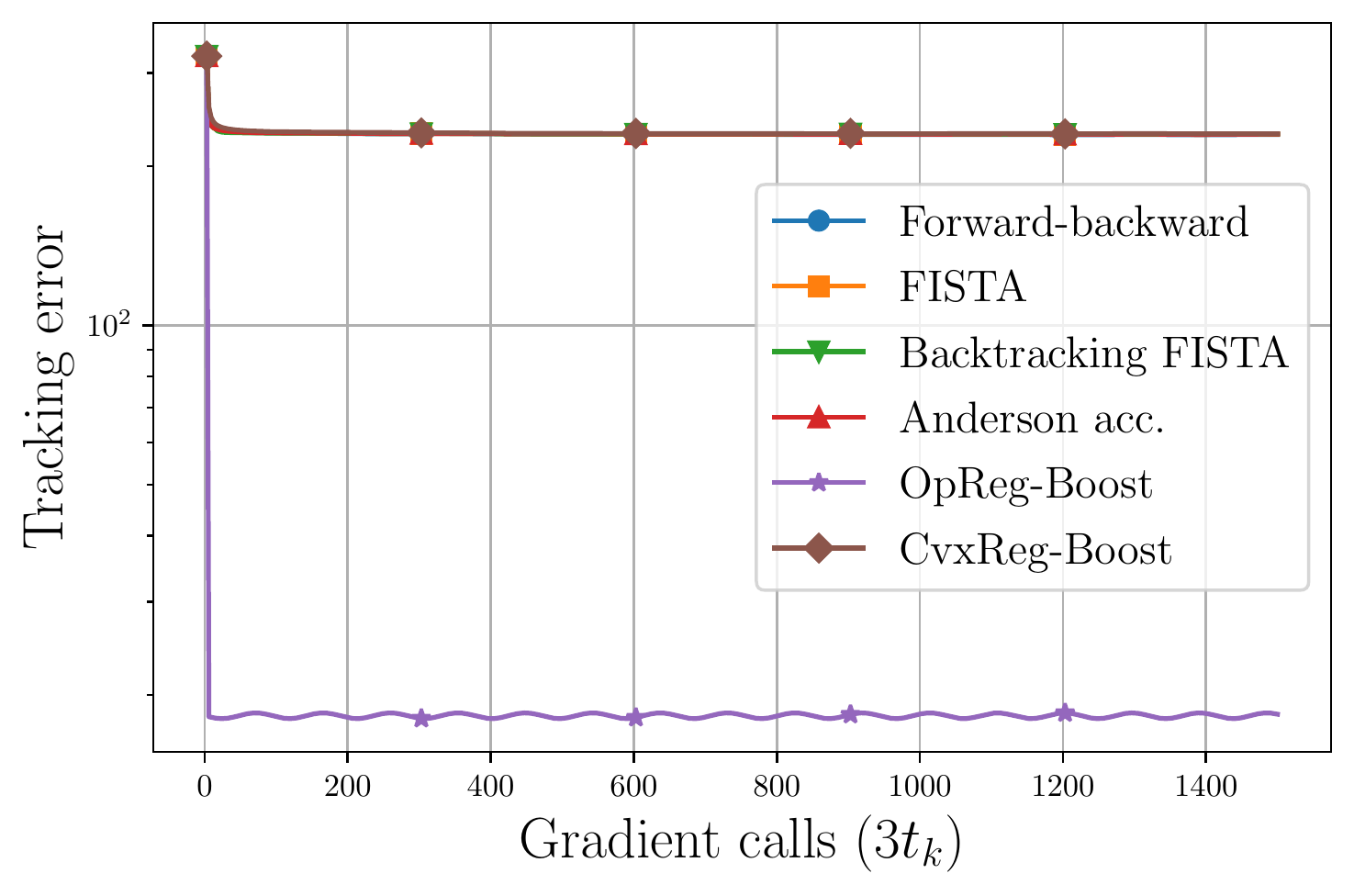}
\caption{Comparison with a fixed computational gradient calls budget ($3$) per time $k \in \N$, with $L = 10^8$ and $\mu = 1$.}
\label{fig:lasso-cvxreg}
\end{figure}

In Figure~\ref{fig:lasso-cvxreg} we report the tracking error for the different methods alongside OpReg-Boost and CvxReg-Boost when applied to the online linear regression problem of section~\ref{subsec:linear-regression}. In particular, all methods are given a budget of $3$ gradient calls per time $k \in \N$, and we choose $L = 10^8$ and $\mu  = 1$. We notice that the performance of CvxReg-Boost in terms of tracking error is very similar to the forward-backward based methods, and it is less performing than OpReg-Boost.

In terms of computational time, OpReg-Boost requires $2.88 \times 10^{-3} s$ per step, whereas for CvxReg-Boost we have $0.36 s$, although both methods have a tailored PRS-based solver. Therefore, the much longer computational time of CvxReg-Boost therefore makes it impractical in the online scenario considered in section~\ref{subsec:linear-regression}.

\section{OpReg-Boost with interpolation}\label{ap.interpolation}
The OpReg-Boost method described in section~\ref{sec:online-opreg} requires that we solve an operator regression problem at each time $k \in \N$. In this section we discuss an alternative approach in which we solve an operator regression problem every $\tau \in \N$ times, and otherwise perform \emph{interpolation} of the solution to the last operator regression problem.

\subsection{Interpolating a Lipschitz continuous operator}
We start by discussing the approach proposed in \cite{valentine_lipschitz_1945} to interpolate Lipschitz continuous operators \emph{while preserving Lipschitz continuity}.

Given the pairs $\{ (\x_i, \hat{\tv}_i) \}_{i \in I_{\ell}}$ -- obtained in our case from the solution of~\eqref{eq:operator-regression} -- we can interpolate the mapping $\hat{\T}_k$ outside the data points as follows. By the Corollary to Theorem~5 of \cite{valentine_lipschitz_1945}, one has that a $\zeta$-Lipschitz continuous map $\hat{\T}_k$ can be extended (that is, interpolated) in a new point $\x$, while preserving Lipschitz continuity (along with its modulus $\zeta$) as follows:
\begin{equation}\label{eq:interpolation}
	\hat{\T}_k(\x) = \begin{cases}
		\hat{\tv}_i & \text{if} \ \x = \x_i \\
		\hat{\tv} \in \bigcap_{i \in I_{\ell}} \ball{\hat{\tv}_i}{\zeta \norm{\x - \x_i}} & \text{otherwise}
	\end{cases}
\end{equation}
where $\ball{\cv}{r} \subset \R^n$ denotes a ball centered at $\cv$ with radius $r > 0$. That is, one can interpolate $\hat{\T}_k$ in a new point $\x$ by finding a point in the intersection of the balls centered in $\hat{\tv}_i$ with radii $\zeta \norm{\x - \x_i}$, $i = 1, \ldots, \ell$. Notice that the intersection $\bigcap_{i \in I_{\ell}} \ball{\hat{\tv}_i}{\zeta \norm{\x - \x_i}}$ is guaranteed to be non-empty by Theorem~1 of \cite{valentine_lipschitz_1945}.

By construction, the new evaluation $\hat{\tv}$ satisfies
$
	\norm{\hat{\tv} - \hat{\tv_i}} \leq \zeta \norm{\x - \x_i}, \forall i \in I_{\ell}, 
$
with $\zeta < 1$, this implies that that contractivity is preserved in this sequential interpolation process.

Interpolating a Lipschitz-continuous operator requires finding a point in the intersection of $\ell$ balls. This can be performed using for example the method of alternating projections (MAP) \cite{reich_projection_2015}. In particular, MAP for operator interpolation based on~\eqref{eq:interpolation} is characterized by the following update:
\begin{equation}\label{eq:alternating_projections}
	\x^{h+1} = (\proj_{\ball{\hat{\tv}_\ell}{\zeta \norm{\x - \x_\ell}}} \circ \cdots \circ \proj_{\ball{\hat{\tv}_1}{\zeta \norm{\x - \x_1}}}) \x^h, \quad h \in \N
\end{equation}
for any initial condition $\x^0 \in \R^n$. In practice we stop MAP when for example $\norm{\x^{h+1} - \x^h} \leq \theta$ for some threshold $\theta > 0$.

\subsection{Interpolation for OpReg-Boost}
In the following we present the interpolated version of OpReg-Boost, in which we keep the estimated operator $\hat{\T}_k$ for a few time instances (termed hereafter as interpolation steps $\tau$), and we interpolate it to generate new approximate optimizers $\x_{k+q}$, $q = 1, \ldots, \tau$. This approach lowers the number of gradient calls, while introducing a lag-type error (since $\hat{\T}_k$ refers to $k$ and will be outdated at $k + \tau$).

\smallskip
\smallskip
\hrule
{\bf OpReg-Boost with interpolation algorithm}
\smallskip
\hrule
{\bf Required:} number of points $\ell$, stepsize $\alpha$, initial condition $\x_0$, interpolation steps $\tau$.

At each time $k$ do:
\begin{itemize}
	\item[\textbf{[S1]}] {\bf IF} $k \,\textrm{mod}\, \tau = 0$ then, 
	\begin{itemize}
	 \item[\textbf{[S2]}] Learn the closest contracting operator to $\T_{k}$, say $\hat{\T}_k$ by employing steps \textbf{[S1.1]}-\textbf{[S1.4]} of OpReg-Boost. Output $\hat{\tv}_k = \hat{\T}_k(\x_{k-1})$.
\end{itemize}
	\hspace{-.4cm} {\bf ELSE}
\begin{itemize}
	\item[\textbf{[S2']}] Interpolate last available $\hat{\T}_k$ for $\tv_k = \x_{k-1} - \alpha \nabla_{\x} f_k(\x_{k-1})$ by using~\eqref{eq:interpolation} with the method of alternating projections and output $\hat{\tv}_k$ 
\end{itemize}
	\item[\textbf{[S3]}] Apply $\x_{k} = \prox_{\alpha g_k} (\hat{\tv}_k) $.
\end{itemize}
\smallskip
\hrule

Finally, in Table~\ref{tab:varying-penalty} we compare the performance of the interpolated version with the standard OpReg-Boost as a function of the PRS penalty parameter $\rho$. We apply the two methods to the online linear regression problem of section~\ref{subsec:linear-regression} when $L = 10^8$ and $\mu = 1$, choosing $\tau = 1$ for the interpolated version -- that is, we solve a new operator regression every other time $k \in \N$.

First of all, it is interesting to notice that for $\rho < 10^{-2}$ the performance of the two versions is almost equal in term of asymptotic error, although as mentioned above the interpolation suffers from an additional lag-type error. However, the time required to carry out the interpolation is larger than the time it takes to solve the operator regression. On the other hand, when we require better precision in the solution of the operator regression by choosing $\rho = 10^{-2}$ we can see that the computational time of OpReg-Boost becomes larger than the interpolation time. This suggests that the choice of applying interpolation has to be made depending on the value of $\rho$.

\begin{table}[!ht]
  \caption{Performance of OpReg-Boost and its interpolated version for different penalty parameters $\rho$.}
  \label{tab:varying-penalty}
  \centering
  \begin{tabular}{lllll}
    \toprule
      &   \multicolumn{2}{c}{OpReg-Boost} &   \multicolumn{2}{c}{OpReg-Boost (interp.)}   \\
    \cmidrule(r){2-3} \cmidrule(r){4-5}
    $\rho$        & As. err.   & Time [s]     & As. err.  & Time interp. [s] \\
    \midrule
    $10^{-6}$     & 18.49      & \textbf{0.00243}      & 18.49      & 0.06361      \\
    $10^{-5}$     & 18.49      & \textbf{0.00243}      & 18.49      & 0.06340      \\
    $10^{-4}$     & 18.49      & \textbf{0.00243}      & 18.49      & 0.06351      \\
    $10^{-3}$     & 15.38      & \textbf{0.00243}      & 15.60      & 0.01839      \\
    $10^{-2}$     & 13.81      & 0.09569      & 90.60      & \textbf{0.00411}      \\
    $10^{-1}$     & 203.60     & 0.02080      & 216.49     & \textbf{0.00368}      \\
    \bottomrule
  \end{tabular}
\end{table}

\section{Proofs of section~\ref{sec:opreg}}\label{ap.proof-prs}

{\bf Proof of Lemma~\ref{lemma.prs}}
We rewrite here Problem~\eqref{eq:equivalent-problem} for the reader ease,
\begin{subequations}\label{eq:equivalent-problem-dummy}
\begin{align}
	&\min_{\tv_{i,e}, \tv_{j,e}} \frac{1}{2 (\ell-1)} \sum_{e \in \mathcal{V}} \norm{\begin{bmatrix} \tv_{i,e} \\ \tv_{j,e} \end{bmatrix} - \begin{bmatrix} \y_i \\ \y_j \end{bmatrix}}^2 \\
	&\text{s.t.} \ \norm{\tv_{i,e} - \tv_{j,e}}^2 \leq \zeta^2 \norm{\x_i - \x_j}^2 \label{eq:interpolation-constraints} \\
	&\qquad \tv_{i,e} = \tv_{i,e'} \ \forall e, e' | i \sim e, e'. \label{eq:consensus-constraints}
\end{align}
\end{subequations}
Let $\xx$ be the vector stacking all the $\tv_{i,e}$, then problem~\eqref{eq:equivalent-problem-dummy} is equivalent to
\begin{equation}
	\min_{\xx} \psi(\xx) + \chi(\xx)
\end{equation}
where
\begin{equation}
	\psi(\xx) = \frac{1}{2(\ell-1)} \norm{\xx - \y}^2 + \iota(\xx)
\end{equation}

with $\iota(\xx) = \sum_{e \in \mathcal{V}} \iota_{e}(\tv_{i,e},\tv_{j,e})$ and $\iota_e$ the indicator function imposing~\eqref{eq:interpolation-constraints}, and $\chi$ the indicator function imposing the ``consensus'' constraints~\eqref{eq:consensus-constraints}. The problem can then be solved using the Peaceman-Rachford splitting (PRS) (see \emph{e.g.} \cite{bauschke_convex_2017}) characterized by the following updates $h \in \mathbb{N}$:
\begin{subequations}
\begin{align}
	&\xx^h = \prox_{\rho \psi}(\z^h) \label{eq:prs-xi} \\
	&\vv^h = \prox_{\rho \chi}(2 \xx^h - \z^h) \\
	&\z^{h+1} = \z^h + \vv^h - \xx^h.
\end{align}
\end{subequations}
The proximal of $\chi$ corresponds to the projection onto the consensus space, and thus can be characterized simply by
\begin{equation}
	\vv_{i,e}^h = \frac{1}{\ell-1} \sum_{e' | i \sim e'} \left( 2 \tv_{i,e'}^h - \z_{e',i}^h\right).
\end{equation}
Regarding the proximal of $\psi$, $\psi$ is separable, in the sense that it can be written as
\begin{equation}
	\psi(\xx) = \sum_{e \in \mathcal{V}} \left[ \frac{1}{2 (\ell-1)} \norm{\begin{bmatrix} \tv_{i,e} \\ \tv_{j,e} \end{bmatrix} - \begin{bmatrix} \y_i \\ \y_j \end{bmatrix}}^2 + \iota_{e}(\tv_{i,e},\tv_{j,e}) \right].
\end{equation}
Therefore, the update~\eqref{eq:prs-xi} can be performed by solving in parallel the problems
\begin{equation}\label{eq:local-updates}
\begin{split}
	(\tv_{i,e},\tv_{j,e}) &= \argmin_{\tv_{i,e},\tv_{j,e}} \left\{ \frac{1}{2 (\ell-1)} \norm{\begin{bmatrix} \tv_{i,e} \\ \tv_{j,e} \end{bmatrix} - \begin{bmatrix} \y_i \\ \y_j \end{bmatrix}}^2 + \frac{1}{2\rho} \norm{\begin{bmatrix} \tv_{i,e} \\ \tv_{j,e} \end{bmatrix} - \z_e^h}^2 \right\}\\
	&\text{s.t.} \quad \norm{\tv_{i,e} - \tv_{j,e}}^2 \leq \zeta^2 \norm{\x_i - \x_j}^2.
\end{split}
\end{equation}
We remark that problems~\eqref{eq:local-updates} are convex QCQPs in $2n$ variables and with one constraint. \hfill $\blacklozenge$

{\bf Proof of Lemma~\ref{lem:1-constraint-qcqp}}
The KKT conditions for~\eqref{eq:stylized-local-update} are
\begin{subequations}
\begin{align}
	&\begin{bmatrix} \tv_i \\ \tv_j \end{bmatrix} - \begin{bmatrix} \w_i \\ \w_j \end{bmatrix} + \lambda \begin{bmatrix} \tv_i - \tv_j \\ \tv_j - \tv_i \end{bmatrix} = 0 \label{eq:kkt-1} \\
	&\lambda \geq 0 \label{eq:kkt-2} \\
	&\lambda \left( \frac{1}{2} \norm{\tv_i - \tv_j}^2 - b \right) = 0 \label{eq:kkt-3} \\
	&\frac{1}{2} \norm{\tv_i - \tv_j}^2 - b \leq 0. \label{eq:kkt-4}
\end{align}
\end{subequations}
Solving~\eqref{eq:kkt-1} for $\tv_i$, $\tv_j$ we get
\begin{equation}\label{eq:closed-form-opreg-t}
\begin{bmatrix} \tv_i^* \\ \tv_j^* \end{bmatrix} = \frac{1}{1 + 2 \lambda^*}
	\left( \begin{bmatrix}
		1 + \lambda^* & \lambda^* \\ \lambda^* & 1 + \lambda^*
	\end{bmatrix} \otimes \Im_n \right) \begin{bmatrix} \w_i \\ \w_j \end{bmatrix},
\end{equation}
and we are left with the need for finding an expression for the Lagrange multiplier.

Assume now that $\lambda > 0$, from~\eqref{eq:kkt-3} this implies that we need to have
$
	\frac{1}{2} \norm{\tv_i - \tv_j}^2 - b = 0
$
and, using~\eqref{eq:closed-form-opreg-t}, this is equivalent to
$$
	\frac{1}{2 (1 + 2\lambda)^2} \norm{\w_i - \w_j}^2 - b = 0.
$$
This is quadratic equation in $\lambda$, with the larger solution being
$$
	\lambda = \frac{1}{2} \left( \frac{\norm{\w_i - \w_j}}{\sqrt{2 b}} - 1 \right);
$$
finally, we impose~\eqref{eq:kkt-2} to guarantee non-negativity of $\lambda$ which yields
$$
\lambda^* = \max\left\{ 0, \frac{1}{2} \left( \frac{\norm{\w_i - \w_j}}{\sqrt{2 b}} - 1 \right) \right\},
$$
and the thesis is proven. \hfill $\blacklozenge$

{\bf Proof of Lemma~\ref{lem:complexity}}
By Lemma~\ref{lemma.prs}, at each iteration the PRS needs to solve the problems~\eqref{eq:prs-local-update}, which are $\ell (\ell-1) / 2$ QCQPs in $2n$ variables and with a single constraint. We can see that these problems are independent of each other, and so that they can be solved in parallel.

As proved in Lemma~\ref{lem:1-constraint-qcqp}, QCQPs with one constraint admit a closed form solution, which has a computational complexity of $O(n)$. The remaining two steps of PRS then require only vectors sums, and are again $O(n)$.

To conclude, at each iteration the complexity is dominated by the need to solve $\ell (\ell-1) / 2$ 1-constraint QCQPs, and so overall we have a complexity of $O(\ell^2 n)$. In the particular case of $n \gg \ell$, which is the typical scenario in practice, then this complexity reduces to $O(n)$. \hfill $\blacklozenge$

\end{document}